%% file: main.tex
\documentclass[conference]{IEEEtran}
\IEEEoverridecommandlockouts
\usepackage{times}

\usepackage[numbers]{natbib}
\usepackage{multicol}
\usepackage[bookmarks=true]{hyperref}

\usepackage{amssymb,amsmath}
\usepackage{graphicx}
\usepackage[ruled, vlined, linesnumbered]{algorithm2e}
\usepackage{url}            
\usepackage{booktabs}       
\usepackage{amsfonts}       
\usepackage{nicefrac}       
\usepackage{microtype}      
\usepackage{xcolor}
\usepackage{amsfonts,amsmath, amsthm, amssymb}
\usepackage{bbm}
\usepackage[ruled,vlined]{algorithm2e}
\DontPrintSemicolon
\usepackage{thmtools, thm-restate}
\usepackage{graphicx}
\usepackage{doi}
\usepackage{bm}
\usepackage{etoc}
\usepackage{hyperref}
\usepackage{tabularx}
\usepackage{amsmath,amssymb,amsthm,tikz,color,chngpage,soul,hyperref,csquotes,graphicx,floatrow, yfonts}
\usepackage{multirow}
\usepackage{subcaption}
\usepackage{tabularx}
\usepackage{mdframed} %
\usepackage{listings}%
\definecolor{light-gray}{gray}{0.95}
\let\labelindent\relax
\usepackage{enumitem}
\usepackage{amsthm}
\usepackage[subtle]{savetrees}

\setlist{nosep}
\renewcommand{\baselinestretch}{0.98}
\usepackage[capitalise,nameinlink]{cleveref}    
\crefformat{equation}{#2(#1)#3}
\crefformat{figure}{#2Fig.~#1#3}
\crefformat{section}{#2\S#1#3}
\crefformat{subsection}{#2\S#1#3}
\crefformat{subsubsection}{#2\S#1#3}
\crefformat{assumption}{#2Assumption~#1#3}
\crefformat{assumption}{#2Assumption~#1#3}

\DeclareMathOperator*{\minimize}{\mathrm{minimize}}
\DeclareMathOperator*{\subjectto}{\mathrm{s.t.}}

\begin{document}
\input{commands}

\title{Real-Time Anomaly Detection and Reactive Planning with Large Language Models}




%
\author{Rohan Sinha\textsuperscript{1}, Amine Elhafsi\textsuperscript{1}, Christopher Agia\textsuperscript{2}, Matthew Foutter\textsuperscript{3}, Edward Schmerling\textsuperscript{4} and Marco Pavone\textsuperscript{1,4}
    \thanks{\textsuperscript{1}Dept. of Aeronautics and Astronautics, Stanford University. \textsuperscript{2}Dept. of Computer Science, Stanford University. \textsuperscript{3}Dept. of Mechanical Engineering, Stanford University. \textsuperscript{4}NVIDIA. Contact: \texttt{\{rhnsinha, amine, cagia, mfoutter, pavone\}@stanford.edu}, \texttt{eschmerling@nvidia.com}.}
}


\maketitle

\begin{abstract}
Foundation models, e.g., large language models (LLMs), trained on internet-scale data possess zero-shot generalization capabilities that make them a promising technology towards detecting and mitigating out-of-distribution failure modes of robotic systems. Fully realizing this promise, however, poses two challenges: (i) mitigating the considerable computational expense of these models such that they may be applied online, and (ii) incorporating their judgement regarding potential anomalies into a safe control framework. In this work, we present a two-stage reasoning framework: First is a fast binary anomaly classifier that analyzes observations in an LLM embedding space, which may trigger a slower fallback selection stage that utilizes the reasoning capabilities of generative LLMs. These stages correspond to branch points in a model predictive control strategy that maintains the joint feasibility of continuing along various fallback plans to account for the slow reasoner's latency as soon as an anomaly is detected, thus ensuring safety. We show that our fast anomaly classifier outperforms autoregressive reasoning with state-of-the-art GPT models, even when instantiated with relatively small language models. This enables our runtime monitor to improve the trustworthiness of dynamic robotic systems, such as quadrotors or autonomous vehicles, under resource and time constraints.
Videos illustrating our approach in both simulation and real-world experiments are available on our project page: \url{https://sites.google.com/view/aesop-llm}.
\end{abstract}

\IEEEpeerreviewmaketitle

\section{Introduction}
Autonomous robotic systems are rapidly advancing in capabilities, seemingly on the cusp of widespread deployment in the real world.
However, a persistent challenge is that the finite datasets used to develop these systems are unlikely to capture the limitless variety of the real world, leading to unexpected failure modes when conditions deviate from training data, or when the robot encounters rare situations that were not well-represented at design time. To mitigate the resulting safety implications, we require methods that can 1) assess the reliability of a machine learning (ML) enabled system at runtime and 2) judiciously enact safety-preserving interventions if necessary.

In this work, we investigate the utility of foundation models (FMs), specifically, large language models (LLMs), towards these two objectives by employing LLMs as runtime monitors tasked with 1) detecting anomalous conditions and 2) reasoning about the appropriate safety-preserving course of action. We do so because
recent work has shown that the internet-scale pretraining data provides FMs with strong zero-shot reasoning capabilities, which has enabled robots to perform complex tasks~\cite{brohan2023can}, identify and correct failures~\cite{huang2022inner}, and reason about potential safety hazards in their surroundings~\cite{elhafsi2023} without explicit training to do so.


However, the adoption of FMs \textit{in-the-loop} of safety-critical robotic systems is immediately met with two challenges.
First, the ever growing scale of FMs poses a major obstacle towards enabling real-time, reactive reasoning about unexpected safety-critical events, especially on agile robotic systems with limited compute. 
Hence, existing work that applies FMs to robotics has focused on quasi-static (e.g., manipulation) or offline settings that afford large times delays while the LLM completes its reasoning.
Second, the application of FMs as runtime monitors requires that they are \textit{grounded} with respect to the task and capabilities of the system.
However, the community has not converged on rigorous methods for grounding FMs without compromising on their generalist zero-shot reasoning abilities (e.g., fine-tuning \cite{KumarRaghunathanEtAl2022} or linear probing \cite{WortsmanIlharcoEtAl2022} often underperform OOD); prompt design remains a standard practice.

\begin{figure}[]
    \centering
    \includegraphics[width=\textwidth]{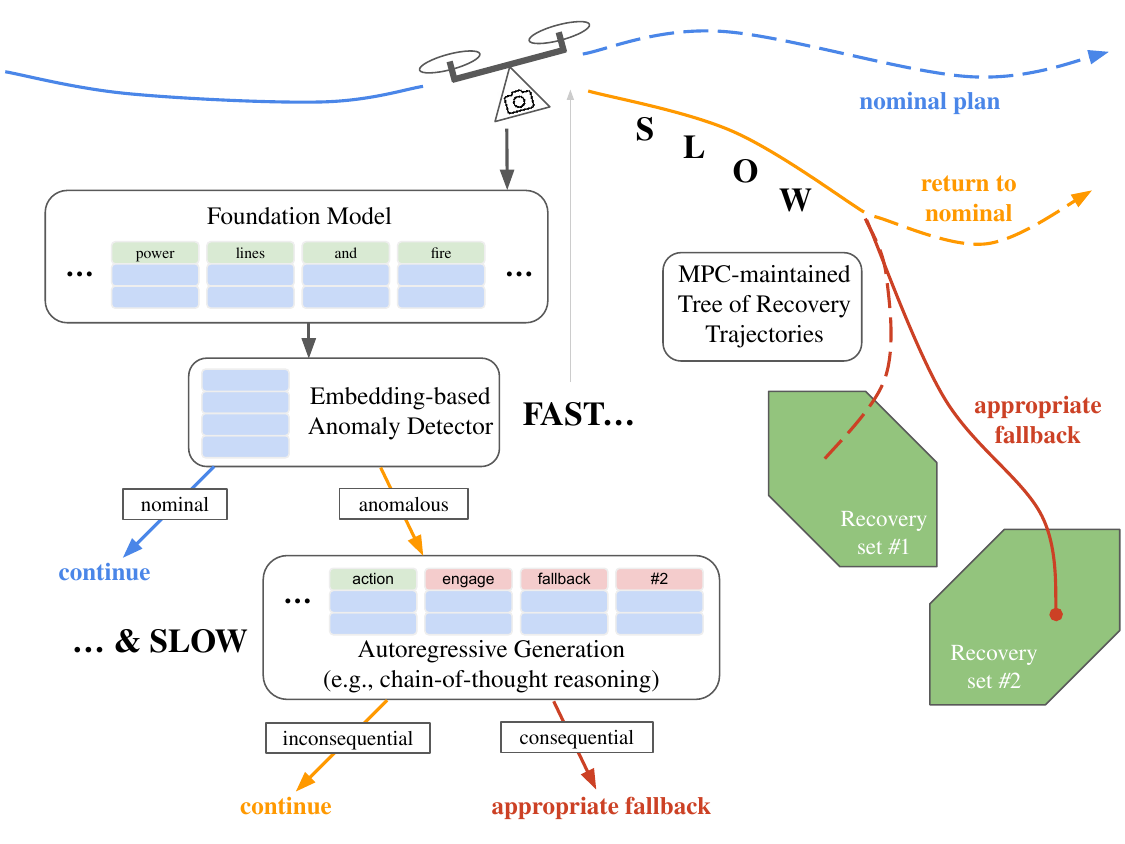}
    \caption{We present an embedding-based runtime monitoring scheme using fast and slow language model reasoners in concert. During nominal operation, \textit{the fast reasoner} differentiates between nominal and anomalous robot observations. If an anomaly is flagged, the system enters a fallback-safe state while \textit{the slow reasoner} determines the anomaly's hazard. In this fallback-safe state, we guarantee access to a set of safe recovery plans (if the anomaly is consequential) and access to continued nominal operation (if the anomaly is inconsequential).}
    \label{fig:aesop-hero}
\end{figure}

To address these challenges, we present AESOP\footnote{This name is inspired by the author of ``The Tortoise and the Hare," in reference to our slow and fast reasoners.}, an anomaly detection and reactive planning framework that aims to derive maximum utility of an LLM's zero-shot reasoning capabilities while taking LLM inference latencies into account within the control design.
As shown in \cref{fig:aesop-hero}, AESOP splits the monitoring task into two separate stages: The first is rapid, real-time detection of anomalies---conditions that deviate from the nominal conditions where the robot performs reliably---by querying similarity with previously recorded observations within the contextual embedding space of an LLM. The second stage is slower, methodical generative reasoning on how to respond to an anomalous scenario once it has been detected. We combine the resultant monitoring pipeline with a model predictive control strategy that maintains multiple trajectory plans, each corresponding to a safety-preserving intervention, in a way that ensures their joint feasibility for an upper bound on the time it takes the slower, generative reasoning to complete\footnote{This approach parallels ideas from dual process theory in cognitive science, popularized in Kahneman's ``Thinking, Fast and Slow''~\cite{kahneman2011thinking}. 
Most of the time, we drive a car based on intuition without careful thought. It is only once something unusual startles us that we carefully reason about how to proceed, often proactively lifting from the throttle to slow down and buy ourselves time to come to a decision.}.
As such, our contributions are threefold:
\begin{enumerate}
    \item \emph{Fast reasoning with embeddings:} We propose a real-time anomaly detection method that, using relatively small FMs (e.g., $120$M parameters) and the robot's previous nominal experiences, surpasses generative chain-of-thought (CoT) reasoning with high-capacity LLMs such as GPT-4. Our method runs at $20\text{Hz}$ on an Nvidia Jetson AGX ORIN, a $357$x speed up over cloud querying GPT-4. To our knowledge, this is the first application of FM embeddings to the task of runtime monitoring, enabling safe and real-time control of an agile robotic system.
    \item \emph{Slow reasoning through autoregressive generation:} While the faster anomaly detector merely detects deviations from prior experiences, we show that autoregressive generation of longer output sequences allows the LLM-based monitor to methodically reason about the safety consequences of out-of-distribution scenarios and decide whether intervention is necessary in a zero-shot fashion; i.e., not all anomalies lead to system-level failures.
    \item \emph{Hierarchical multi-contingency planning:} Facilitated by our fast anomaly detector, we introduce a predictive control framework to integrate both FM-based reasoners in a lower-level reactive control loop by maintaining multiple feasible trajectories, each corresponding to a high-level intervention strategy. This allows the robot to 1) react to sudden semantic changes in the robot's environment, 2) maintain closed-loop safety while waiting for a slow reasoner to return a decision, and 3) exhibit dynamic, agile behaviors within the range of scenarios where the nominal autonomy stack is trustworthy.
\end{enumerate}
We demonstrate these facts across several commonplace LLMs, ranging from $10^8-10^{12}$ parameters, as well as conventional OOD detection techniques on 1) an extensive suite of synthetic text-based domains, 2) \textbf{simulated and real-world} closed-loop quadrotor experiments resembling a drone delivery service, and 3) careful recreations of recent real-world failure modes of autonomous vehicles in the CARLA simulator~\cite{dosovitskiy2017carla}. 
We conclude that the use of FMs not only presents a promising direction to significantly improve the robustness of autonomous robotic systems to out-of-distribution scenarios, but also that their real-time integration within dynamic, agile robotic systems is already practically feasible.


\textbf{Organization:} We first discuss related work in \cref{sec:relwork} and formalize the problem setup in \cref{sec:problem-formulation}. Then, we present our approach in \cref{sec:approach} and evaluate our method in \cref{sec:expts}. Finally, we conclude and provide a future outlook in \cref{sec:outlook}. In addition, we include a full overview of the notation and conventions used in this paper in \cref{ap:glos}.
\section{Related Work}\label{sec:relwork}
\textbf{Out-of-Distribution Robustness:} The fact that learning-based systems often behave unreliably on data that is dissimilar from their training data has been extensively documented in both the machine learning and robotics literature \cite{GeirhosJacobsenEtAl2020, RechtRoelofsEtAl2019, OvadiaFertigEtAl2019, SinhaSharmaEtAl2022}. Approaches to address the subsequent challenges broadly fall into two categories \cite{SinhaSharmaEtAl2022}: First are methods that strengthen a model's performance in the face of distributional shift. For example, through robust training (e.g., \cite{SagawaKohEtAl2020}) or by adapting the model to changing conditions (e.g., \cite{HoffmanTzengEtAl2018, chen2023adapt}). Second are so called out-of-distribution detection algorithms \cite{SalehiMirzaeiEtAl2021, RuffKauffmanEtAl2021}, that aim to detect when a given model is unreliable, e.g., by computing the variance of an ensemble \cite{LakshminarayananPritzelEtAll2017} or computing energy scores \cite{liu2020energy}. Recent work has shown the merits of generalist FMs like LLMs in both domains: Studies have shown that zero-shot application of a FM (e.g., in \cite{WortsmanIlharcoEtAl2022}, the authors apply CLIP zero-shot on ImageNet), vastly improves OOD generalization over previous approaches, like distributionally robust training \cite{MillerTaoriEtAl2021, WortsmanIlharcoEtAl2022, brown2020language}. 
In addition, existing OOD detection methodologies and their application within robot autonomy stacks are tailored to detect conditions that compromise the reliability of individual components of an autonomy stack, like whether a perception system's detections are correct \cite{RichterRoy2017, FilosTigas2020, RahmanCorkeEtAl2021, SinhaSchmerlingEtAl2023}. Instead, recent work showed that LLMs may provide a more general mechanism to detect context dependent safety hazards, especially those that are hard to measure with predefined performance metrics \cite{elhafsi2023}. For example, an autonomous EVTOL may monitor the quality of the vision system's landing pad location estimate, but even if the EVTOL has high confidence that it can land successfully, the outcome of landing on a building that is on fire can have profound negative consequences. However, despite the attractive properties of LLMs, these works do not propose practical strategies to integrate them in closed-loop. Therefore, we propose a closed-loop control framework that can both use the LLM to identify unseen anomalies and strengthen performance in the presence of rare failure modes.
    \textbf{Foundation Models in Robotics}:
    The integration of large lanugage models (LLMs) and, more broadly, foundation models (FMs) into robotics has sparked considerable interest due to their proficiency in managing complex, unstructured tasks that demand sophisticated reasoning skills. These models have been instrumental in bridging the gap between natural language instructions and the execution of physical actions in the real world. Various approaches utilizing these models have been developed for online use in applications in areas such as manipulation~\cite{huang2023voxposer}, navigation~\cite{shah2023lm}, drone flight~\cite{chen2023typefly}, and long-horizon planning~\cite{brohan2023can,Lin2023}. FMs have also been used to define reinforcement learning reward functions~\cite{yu2023language}, generate robot policy code~\cite{liang2023code}, or create additional training data~\cite{xiao2022robotic, yu2023scaling, ahn2024autort}.

    However, the issue of response time associated with FMs has not been a focal point in these studies. The aforementioned online methods predominantly rely on a quasi-static assumption, implying that the timing of the robot's actions is not critical. This assumption allows the system the luxury of time to consult the LLMs and await their responses without urgency. Conversely, the latter methods either operate offline or utilize LLMs in manners that are similarly insensitive to response time. 
    
    As such, existing work demonstrates limited dynamic reactivity of the policy, which is essential for fast-moving, agile robots like quadrotors. These robots can quickly find themselves in situations where a delayed response can result in an unavoidable crash. To mitigate this issue, our approach specifically considers the delays introduced by the reasoning process. We enhance reactivity by implementing a more rapid anomaly detection system, thereby reducing the risk of crashes by allowing for timely corrective actions.

    \textbf{Accelerating Inference:} It is well-recognized that increasing FM capabilities are accompanied with increasing computational cost and inference latency. As such, substantial effort is being dedicated to the acceleration of these models, of which several popular strategies have emerged such as model distillation~\cite{hinton2015distilling, hsieh2023distilling}, quantization~\cite{jacob2018quantization, xiao2023smoothquant}, and parameter sparsification~\cite{sun2023simple, panigrahi2023task}. 
    Ultimately, these approaches improve the cost of the forward pass through a transformer model, but do not address the fact that LLMs typically need to generate long sequences of outputs to reason towards the correct decision~\cite{wei2022chain}, a process unlikely to run in real-time for time-sensitive tasks. Querying remotely hosted models on large-scale hardware (e.g., GPT-4 \cite{achiam2023gpt}) is a potential solution if computation constraints become too stringent for a system to perform onboard FM inference, yet network conditions may incur inconsistent and potentially significant delays, and connectivity may be unreliable for in-the-wild deployments. 

\section{Problem Formulation}\label{sec:problem-formulation}
In this work, we consider a robot with discrete time dynamics
\begin{equation} \label{eq:dyn}
    \x_{t+1} = \f(\x_t, \fu_t),
\end{equation}
where $\x_t \in \R^n$ represents the robot's state, and $\fu_t \in \R^m$ is the control input. Nominally, we aim to minimize some control objective $C$ that depends on the states and inputs, subject to safety constraints on the state $\x_t\in\calX\subseteq \R^n$ and input $\fu_t\in\calU \subseteq \R^m$. For example, a quadrotor's state consists of its pose and velocity (estimated from e.g., GPS, visual-SLAM, and IMUs), and its objective may be to minimize distance to a landing zone subject to collision avoidance constraints. 

In addition to the state variables tracked by the nominal control loop, the robot receives an observation $\obs_t \in \calO$ at each timestep, which provides further contextual information about the robot's environment. 
Our goal is to design a runtime monitor that interferes with the nominal system to avoid system-level safety hazards, which may depend on environmental factors not represented in the robot's state $\x_t$. 
For example, a quadrotor cannot safely land on a landing zone covered in burning debris even if the nominal control stack has the ability to do so. 
In the spirit of~\cite{elhafsi2023}, we refer to such events as \emph{semantic failure modes}, as they do not necessarily constitute violation of precise state constraints $\calX$, but instead depend on the qualitative context of the robot's task. 

Further, we assume that we have access to a dataset $\calD_{\mathrm{nom}} = \{\obs_i\}_{i=1}^N$ of nominal observations wherein the robot was safe and reliable. Conceptually, $\calD_{\mathrm{nom}}$ corresponds to the operational data of a notionally mature system, and may consist of data used to train the system or of previously collected deployment data. This data will overwhelmingly contain mundane scenarios where the robot performs well; our monitoring framework is targeted instead at the challenging, extremely rare corner cases that are unlikely to have been recorded before and threaten the robot's reliability.

In the event that a failure mode of the nominal autonomy stack is imminent, we must select and engage a safety-preserving intervention. For example, we may choose to land the quadrotor in another open landing zone like a grassy field. To this end, we follow \cite{SinhaSchmerlingEtAl2023} and we assume that we are given a number of \emph{recovery regions} $\calX_R^1, \calX_R^2, \dots, \calX_R^d \subseteq \calX$, control invariant subsets of the state space that correspond to high-level safety interventions. For example, $\calX_R^i$ may represent the alternate landing zone. By planning a trajectory to the appropriate recovery set, potential safety hazards can be avoided. As shown in \cite{SinhaSchmerlingEtAl2023}, such sets can both be hand defined up-front and identified using reachability analysis. 

\section{Proposed Approach}\label{sec:approach}
It is virtually impossible to account for all the corner-cases and semantic failure modes that a system may experience through a standard engineering pipeline.
Even if we train e.g., classifiers to detect obstructions on landing zones, there may always remain a class of semantic failure modes that we have not accounted for. Instead, we propose to leverage generalist foundation models to detect and reason holistically about a robot's environment. We first present our FM-based monitoring approach, after which we construct a planning algorithm that accounts for the latency that FM-based reasoning may induce. 
\subsection{Runtime Monitor: Fast and Slow Reasoning}\label{sec:fast-and-slow}
To detect and avoid semantic failure modes, we propose a two-stage pipeline. The first is the detection of anomalies, simply defined as conditions that deviate from the mundane, nominal experiences where we know that our notionally mature system is reliable. The second is slower reasoning about the downstream consequence of an anomaly, if detected, towards a high-level decision on whether a safety-preserving intervention should be executed.
We refer to \cref{sec:prelim-anomaly} for a brief introduction to anomaly detection used hereafter.

\textbf{Fast Anomaly Detection:} To detect anomalies, we need to inform a FM of the context within which the autonomous system is known to be trustworthy.
The prior, nominal experiences of the robot serve as such grounding. We construct an anomaly score function $\s(\obs_t, \calD_\mathrm{nom}) \in \R$ to query whether a current observation $\obs_t$ differs from the previous experiences in $\calD_{\mathrm{nom}}$. We do not require any particular methodology to generate the score, we just require that scoring an observation is computationally feasible in real-time; that is, within a single time step. 

This work emphasizes the value of computing anomaly scores using language-based representations, which we show capture the semantics of the observation within the context of the robot's task in \cref{sec:expts}. 
To do so, we first create a cache of embedding vectors $\calD_e = \{\e_i\}_{i=1}^N$ where $\e_i = \phi(\obs_i) \in \R^e$ for each $\obs_i \in \calD_{\mathrm{nom}}$ by embedding the robot's prior experiences offline using an embedding FM $\phi$. Then, at runtime, we observe $\obs_t$, compute its corresponding embedding $\e_t$, and compute an anomaly score $\s(\e_t ; \calD_e)$ using the vector cache. We investigate several simple score functions (see \cref{supp:score-fn-analysis} for a full list), each of which roughly measures a heuristic notion of difference with respect to $\calD_{\mathrm{nom}}$. For example, the simplest metric uses the maximum cosine similarity with respect to samples in the prior experience cache, 
\begin{equation*}
    \s(\e_t; \calD_e) := - \max_{\e_i \in \calD_{e}} {\frac{\e_i^T \e_t}{\|\e_i\|\|\e_t\|}},
\end{equation*}
 which, in effect, retrieves the most similar prior experience from $\calD_e$ to construct the score. Intuitively, this approach measures whether anything similar to the current observation has been seen before.


Finally, to classify whether an observation should be treated as nominal or anomalous, we can calibrate a threshold $\tau \in \R$ as the $\alpha \in (0,1)$ quantile of the nominal prior experiences, 
\begin{equation}\label{eq:emp-quantile}
        \tau = \inf\bigg\{q\in \R\ : \  \frac{|{\e_i \in \calD_e: \s(\e_i; \calD_e \setminus \{\e_i\})\leq q}|}{N} \geq \alpha\bigg\},
\end{equation}
i.e., the smallest value of $q$ that upper bounds at least $\alpha N$ nominal samples. Note that for nominal embeddings, we must compute the anomaly score $\s$ in a leave-one-out fashion, since $\s(\e_i ; \calD_e) = -1$ for $\e_i \in \calD_e$. Determining the threshold $\tau$ using empirical quantiles as in \cref{eq:emp-quantile} is a standard approach \cite{RichterRoy2017}, but could be extended in future work to make precise guarantees on false positive or negative rates using recent results in conformal prediction~\cite{AngelopoulosBates2022, LuoZhaoEtAl2021}.


\textbf{Slow Generative Reasoning:}
Once we detect an anomaly, we trigger the autoregressive generation of an LLM to generate a zero-shot assessment of whether we need to engage any of the interventions associated with the recovery sets $\calX_R^1, \dots, \calX_R^d$ (\cref{sec:problem-formulation}) to maintain the safety of the system.
The value of this approach is that the LLM's internet-scale pretraining data allows it to generate outputs that resemble the generalist common sense reasoning that a human operator is likely to suggest, as a result, making superior decisions on OOD examples, on which existing task-specific learning algorithms are notoriously unreliable. 

To do so, we follow \cite{elhafsi2023} in using a VLM to convert the robot's current visual observation into a text description of the environment. We simply encode this scene description into a prompt that provides context on the monitoring task, as illustrated in \cref{fig:aesop-drone}. We then parse the resulting output string to yield a classification $y \in \{0, 1, \dots d\}$ on whether the anomaly does not present a hazard and the system can continue it's nominal operation ($y=0$), or whether we should engage intervention $y \in \{1, \dots, d\}$ and steer the state into recovery set $\calX_R^y$. As we illustrate in our experiments, the recovery sets naturally correspond to high-level behaviors (e.g., landing in a field), which facilitates prompt design.
We use the shorthand $\w(\obs_t, \calY)$ to denote the output of the slow reasoner when given observation $\obs_t$ and a (sub)set of intervention strategies $\calY \subseteq \{1, \dots, d\}$.



Whether inference is run onboard or the model is queried remotely over unreliable networks in the cloud, we must account for the latency that autoregressive reasoning introduces. 
For example, a fast moving vehicle may collide with an anomalous obstacle if it's reaction time is too slow. 
Therefore, we account for the LLM's compute latency by assuming that it takes at most $K \in \mathbb{N}_{>0}$ timesteps to receive the output string from the slow reasoner. It is usually straightforward to identify the value of $K$ in practice, since we prompt the model to adhere to a strict output template that tends to stabilize the length of the output generations. Alternatively, as we describe in \cref{sec:exps-hardware} and \cref{app:hardware}, a simple field-test can be sufficient to identify an upper bound on typical network latency.

\subsection{Planning a Tree of Recovery Trajectories}\label{sec:fallback-planning}
We control the robot's dynamics \cref{eq:dyn} in state-feedback using a receding horizon control strategy that 1) minimizes the nominal control objective along a horizon of $T > K$ timesteps, while 2) maintaining a set of $d$ recovery trajectories that each reach one of the respective recovery sets $\calX_R^i$ within the horizon $T$. The goal of this approach is to ensure that the high-level safety interventions provided to the slow reasoner can be executed. Additionally, it is essential that these options remain feasible throughout the $K$ time steps it takes the monitor to decide on the most appropriate choice. Otherwise, a fast moving robot may, for example, no longer be able to stop in time to avoid a collision. To this end, we solve the following finite-time optimal control problem online, which maintains a consensus between the recovery trajectories for $K$ timesteps:
\begin{align}\label{eq:modification-mpc}
&J_t(\calY, K, T) = \nonumber\\
&\begin{aligned}
    \minimize_{ \{\x_{t:t+T+1|t}^i, \fu_{t:t+T|t}^i\}_{i\in\calY \cup \{0\}}} 
    &\enspace C(\x_{t:t+T+1|t}^0, \fu_{t:t+T|t}^0)\\
    \subjectto\enspace & \x_{t+k+1|t}^i = \f(\x_{t+k|t}^i, \fu_{t+k|t}^i) \\
    & \fu_{t+k|t}^i \in \calU \quad \x_{t+k|t}^i \in \calX \\
    & \x_{t|t}^i = \x_t \\
    & \x_{t+T+1|t}^i \in \calX_R^i  \quad \forall i \in \calY \\
    & \fu_{t|t}^i = \fu_{t|t}^0 \quad \forall i \in \calY \\
    & \fu_{t:t+K|t}^i = \fu_{t:t+K|t}^j \  \forall i, j \in \calY
\end{aligned}
\end{align}
Here, the notation $\x^i_{t+k|t}$ indicates the predicted value of variable $\x$ at time $t+k$ computed at time $t$ for each trajectory $i \in \calY \cup \{0\}$.
The MPC in \cref{eq:modification-mpc} optimizes a set of $|\calY| + 1$ trajectories. The first corresponds to a nominal trajectory $\x_{t:t+T+1|t}^0$ plan that minimizes the control objective and a set of $|\calY|$ recovery trajectories that each reach their respective recovery set $\calX_R^i$ within $T$ timesteps. In addition, the MPC problem \cref{eq:modification-mpc} includes two consensus constraints, one associated with the fast anomaly detector and the other with the slow reasoner. First, by fixing consensus along the first input of the nominal trajectory and all the recovery trajectories, we ensure that the set of feasible interventions is non-empty during nominal operation. The second fixes consensus for $K$ timesteps along the set of recovery trajectories, in effect generating a branching tree of recovery trajectories. If we then use the fast anomaly detector to both trigger execution of the first $K$ actions of the recovery trajectories and the slower reasoning, we ensure that the options we provide to the slow reasoner are still available when it returns its output. 

We summarize this methodology in Algorithm \ref{alg:safe-mpc}, which guarantees that we reach the recovery set chosen by the slow reasoner within at most $T+1$ timesteps after detecting an anomaly:

\begin{algorithm}[t]
\caption{AESOP}
\label{alg:safe-mpc}
  \SetAlgoLined
  \KwIn{State $\x_0$ such that \cref{eq:modification-mpc} is feasible, fast anomaly detector $\h$, slow reasoner $\w$ with latency $\leq K$. 
  
  }
  $t_{\mathrm{anom}} \gets \emptyset$

  \For{ $t = 0, 1, 2, \dots$}{
    Observe $\x_t, \obs_t$
    
    \If{$t_{\mathrm{anom}} = \emptyset$ or $\w(\obs_{t_{\mathrm{anom}}}, \calY_{t_{\mathrm{anom}}}) = 0$}{ 
   
    Choose $\calY_t \subset \{1, \dots d\}$ s.t. \cref{eq:modification-mpc} is feasible with arguments $\calY_t$, consensus horizon $K$

    Solve \cref{eq:modification-mpc} with $J_t(\calY_t, K, T)$

    \If {$\h(\obs_t) = \mathrm{True}$} {
       $t_{\mathrm{anom}} \gets t$

    $\w(\obs_t, \calY_t)\mathrm{.start()}$
    }
    
    Apply optimal control input $\fu^{\star, 0}_{t|t}$
    }
    \ElseIf {$t_{\mathrm{anom}} \neq \emptyset$ and not $\w(\obs_{t_{\mathrm{anom}}}, \calY_{t_{\mathrm{anom}}}){\mathrm{.done()}}$}{
    
    Set $k \gets t - t_{\mathrm{anom}}$ 
    
    Solve \cref{eq:modification-mpc} with $J_t(\calY_{t_{\mathrm{anom}}}, K - k, T - k)$
    
    Apply optimal control input $\fu^{\star, 0}_{t|t}$
    
    }
    \Else {
    Set $k \gets t - t_{\mathrm{anom}}$ 
    
    Solve \cref{eq:modification-mpc} with $J_t(\{\w(\obs_{t_{\mathrm{anom}}}, \calY_{t_{\mathrm{anom}}})\}, 0, \max\{0, T - k\})$
    
    Apply optimal control input $\fu^{\star, y}_{t|t}$, where $y = \w(\obs_{t_{\mathrm{anom}}}, \calY_{t_{\mathrm{anom}}})$
    }
    }
\end{algorithm}

\begin{restatable}{theorem}{safety}\label{thm:safety}
Suppose that at $t=0$, the MPC in \cref{eq:modification-mpc} is feasible for some set of recovery strategies $\calY \subset \{1, \dots, d\}$, i.e., that $J_0(\calY, K, T) < \infty$. Then, the closed-loop system formed by \cref{eq:dyn} and Algorithm \ref{alg:safe-mpc} ensures the following: 1) We satisfy state and input constraints $\x_t\in\calX$, $\fu_t \in \calU$ for all $t\geq 0$. 2) At any time $t\geq0$, there always exists at least one safety intervention $y \in \{1, \dots, d\}$ for which the MPC \cref{eq:modification-mpc} is feasible. 3) If the slow reasoner $\w$, triggered at some time $t_{\mathrm{anom}}>0$,  chooses an intervention $y\in \{1, \dots, d\}$, then for all $t \geq t_{\mathrm{anom}} + T + 1$ it holds that $\x_t \in \calX_R^y$.
\end{restatable}
\begin{proof}
    See \cref{supp:theorem-proof}.
\end{proof}
    
The emergent behavior of Algorithm \ref{alg:safe-mpc} is that once the fast anomaly detector issues a warning regarding an unusual observation, the robot will balance progress along the nominal trajectory and jointly maintaining dynamic feasibility of the fallback options available at $t_{\mathrm{anom}}$. This generally leads the robot to slow down to preserve its options, thereby providing the slow reasoner with time to think. Upon reaching a decision from the slow reasoner, the robot either transitions back to nominal operations, if the observation is not hazardous, or engages the selected safety intervention and commits the robot to the associated recovery set. While this scheme does not explicitly ensure that multiple strategies remain available to the robot throughout nominal operation (though at least one will remain so), we find in a practical setting (see \cref{sec:exps-drone}, \cref{sec:exps-hardware}) that multiple simple hand-designed intervention strategies remain persistently feasible. Still, methods for dynamically identifying and selecting recovery regions present an exciting avenue for future work.

\section{Experiments}\label{sec:expts}

Having outlined our approach, we conduct a series of experiments to test the following five hypothesis:
\begin{description}
    \item[\textbf{H1}] By quantifying semantic differences of observations with respect to the prior experience of a system, our fast embedding-based anomaly detector performs favorably to generative reasoning-based approaches.
    \item[\textbf{H2}] Embedding-based anomaly detection does not necessitate the use of high-capacity generative models; small models incurring marginal costs can be used.
    \item[\textbf{H3}] Once an \textit{anomaly} is detected, generative reasoning approaches can effectively deduce whether the anomaly warrants enacting safety-preserving interventions.
    \item[\textbf{H4}] Our full approach, which unifies embedding-based anomaly detection and generative reasoning-based anomaly assessment, can be integrated in a broader robotics stack for real-time control of an agile system.
    \item[\textbf{H5}] Additional forms of embeddings, including those from vision and multi-modal models, offer a promising future avenue for end-to-end anomaly detection. 
\end{description}
\textbf{Experiment Rationale:} We run four main experiments.
The first experiment (\cref{sec:exps-synthetics}) tests the performance of our fast anomaly detector in three synthetic (i.e., text-based) robotic environments. 
We then evaluate the slow generative reasoner for the assessment of detected anomalies on two of these environments.
The second experiment (\cref{sec:exps-drone}) evaluates our full approach  (integrating the runtime monitor with the MPC fallback planner) in a simulation of real-time control of an agile drone system. The third is a full-stack experiment on real quadrotor hardware, including a timing breakdown for each component in our approach running on a Jetson AGX Orin module, thereby demonstrating viability for hardware deployment.
The fourth experiment evaluates whether our runtime monitor transfers to a realistic, semantically rich self-driving environment, where we investigate the use of both language and multi-modal embeddings for anomaly detection. 

All code used in our experiments, including scripts to generate the synthetic datasets and prompt templates, can be found through our project page at \url{https://sites.google.com/view/aesop-llm}. In addition, we provide a brief description of our prompting strategy in \cref{ap:prompting}.

\subsection{Synthetics---Manipulation, Autonomous Vehicles, VTOL}\label{sec:exps-synthetics}


We construct three synthetic domains to support our analysis: a Warehouse Manipulator domain, an Autonomous Vehicle domain, and a Vertical Take-off and Landing (VTOL) Aircraft domain. 
Each domain consists of scenarios in which a notionally mature autonomous robot may (or may not) encounter a safety concerning observation during its typical operations.
The robots' observations take the form of a collection of \textit{concepts}, which we define as one or more objects and their semantic relationships.
For example, in the VTOL domain, both ``ice'' and ``helipad'' represent concepts of a single object class, whereas ``icy helipad'' represents a third concept formed by their conjunction.
We follow definition of \textit{anomalies} given in \cref{sec:fast-and-slow}: observations that, in the context of a given task, deviate from the robots' nominal experiences.
Thus, anomalies may not pose safety risks but must still be identified for further analysis. 

\begin{figure*}
    \centering
    \includegraphics[width=0.95\textwidth]{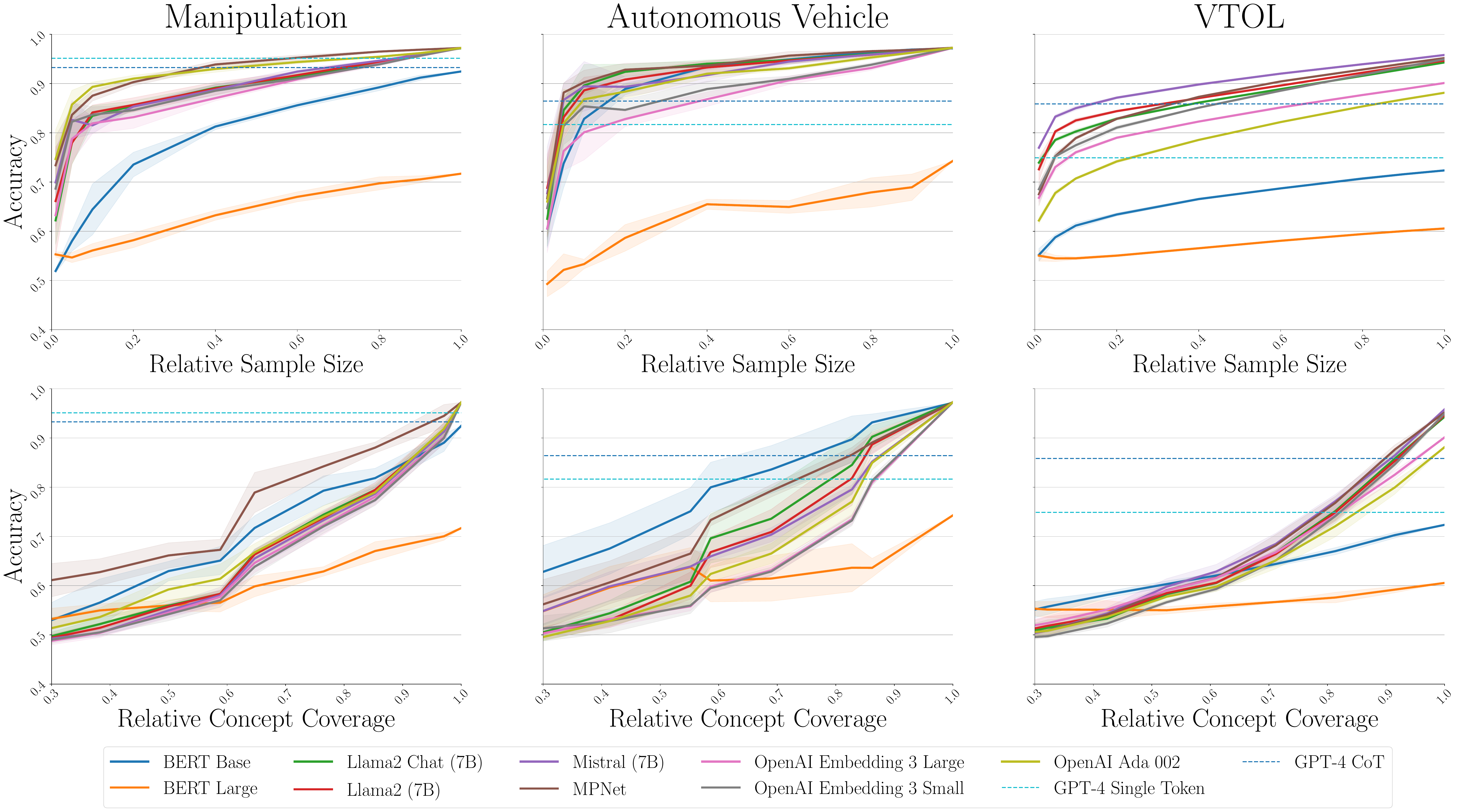}\\
    \caption{Embedding-based (fast) anomaly detection results for the manipulation, autonomous vehicle, and VTOL domains. The top row of figures plot anomaly detection accuracy as a function of experiences sampled IID from the respective domain datasets. The bottom row of figures plot accuracy as a function of the concepts sampled from the respective domain datasets. We use top-5 scoring with the anomaly detection threshold set at the $95$-th quantile \cref{eq:emp-quantile} of the scores in the sampled data.}
    \label{fig:synthetic-results}
\end{figure*}

We briefly describe the synthetic domains below:
\begin{itemize}
    \item \textbf{Warehouse Manipulator (WM):} A mobile WM robot performs the task of ``sorting objects on a conveyor belt.'' Observations contain concepts sampled from a predefined set of conveyor belt (e.g., a package) and surrounding environment (e.g., a storage shelf) objects. Anomalies consist of hazardous objects on the conveyor belt (e.g., a leaking bleach bottle) or object combinations in the surrounding environment (e.g., two forklifts in collision). The dataset contains $1551$ scenarios.
    \item \textbf{Autonomous Vehicle (AV):} An AV operating as a taxi service performs the task of ``driving to a set destination.'' Observations contain concepts sampled from a predefined set of task-relevant (e.g., a car, bus, or traffic light) and task-irrelevant (e.g., an airplane in the sky) objects. Anomalies consist of unusual task-relevant objects (e.g., a blank speed limit sign) or combinations (e.g., a traffic light on a truck). The dataset contains $840$ scenarios.
    \item \textbf{Vertical Take-off and Landing (VTOL):} A VTOL aircraft operating as an urban air taxi performs one of two tasks: ``flying toward a set destination'' or ``landing on a designated building.'' Observations contain concepts sampled from a predefined set of flying objects, landing zones, and ground regions. Anomalies consist of unanticipated flying objects (e.g., a large swarming flock of birds) and/or landing zones (e.g., a building rooftop on fire). The dataset contains $18400$ scenarios.
\end{itemize}
The synthetic domains vary in terms of size and complexity, with WM being the simplest, and VTOL being the most challenging.
Complexity is determined by extent to which anomalous concepts differ from the nominal experiences of the robot \footnote{In the WM domain, anomalous observations consist of overtly abnormal concepts (e.g., a ``smoking lithium battery,'' ``a broken glass bottle'') relative to the typical ``package'' or ``computer'' one would expect on a conveyor belt.
This in turn simplifies the task of differentiating anomalous from nominal observations.
By contrast, the VTOL domain consist of more nuanced concept shifts; an observation containing a ``flying bird'' is conceptually similar to a ``swarming flock of birds,'' though one is nominal, and the other anomalous.}. 

\subsubsection{Fast Reasoning for Anomaly Detection \normalfont{(\textbf{H1}, \textbf{H2})}}\label{sec:exps-fast-reasoning}
The role of the anomaly detector is to analyze the robot's observations and identify whether, due to the presence of an atypical object or concept, an observation qualifies as an anomalous.

\textbf{Methods:} 
We evaluate our fast anomaly detector with nine language models, varying in size and function: BERT-base (110M) and BERT-large (336M) uncased~\cite{devlin2018bert}, Sentence Transformer MPNet (110M; BERT-base architecture trained for embeddings)~\cite{song2020mpnet, reimers-2019-sentence-bert}, completion and instruction-tuned Llama 2 models (7B)~\cite{touvron2023llama}, Mistral (7.11B)~\cite{jiang2023mistral, wang2023improving}, and three OpenAI embedding models (parameters not disclosed). 
The choice of score function $\s(\e_t; \calD_e)$ (e.g., cosine similarity, top-$k$ scoring, Mahalanobis distance) did not yield significant performance variation. Thus, we only report results for top-$5$ scoring and refer to \cref{supp:synthetics} for extended results.

\textbf{Baselines:} Our baselines consist of generative reasoning with GPT-4, queried to classify the robot's observation as ``nominal'' or ``anomalous.'' 
We consider a two variants: single-token (ST) prediction and chain-of-thought reasoning (CoT).
For CoT, GPT-4 first reasons over each \textit{concept} in the observation before outputting an overall anomaly classification.

We use prompts with identical prefixes across all methods, making minor modifications to, e.g., elicit CoT reasoning.
For metrics we report accuracy, setting the detection threshold at the $95$-th quantile (\cref{eq:emp-quantile}) of the scores in the nominal dataset. 

\textbf{Results \& Analysis:} The results are shown in \cref{fig:synthetic-results}.
The first row shows the accuracy of our fast anomaly detectors with increasing sample size (i.e., the size $N$ of the embedding cache $\calD_e = \{\e_i\}_{i=1}^N$) drawn IID from the full dataset of nominal observations $\calD_{\mathrm{nom}}$.
The second row is identical to the first, except, showing accuracy with increasing percent of \textit{concept coverage}; that is, the percent of nominal concepts contained in the embedding cache $\calD_e$ used to construct our detector.

Comparing the top and bottom rows for each domain, we observe that performance increases logarithmically with relative sample size, but only linearly with increasing concept coverage.
This indicates that our fast anomaly detector scales favorably with the diversity of nominal concepts represented in the robot's experience as opposed to the shear scale of experience.
Provided with sufficient concept coverage, our anomaly detector clearly outperforms the single-token and CoT generative reasoning baselines with GPT-4, validating our first hypothesis \textbf{H1}.

Comparing among language models, we find that the performance of models does not strictly correlate with their size, but rather, depends on their training data and strategy.
For example, Sentence Transformer MPNet, a 110M parameter model trained for embeddings, often outperforms BERT Large (336M) and the OpenAI embedding models, and even performs comparatively to the Llama 2 (7B) and Mistral (7.11B) models. This validates our second hypothesis \textbf{H2}.


The key advantage of our anomaly detectors is that they ground the analysis of observations in the embedding space of previously observed concepts.
It is unreasonable to attempt to ground the generative baselines in such a way due to their limited context windows, among other challenges (e.g., recency bias~\cite{zhao2021calibrate}).
Moreover, we see that GPT-4's performance gradually decreases as the complexity of the environments increase (e.g., from AV to VTOL), and as such, we may expect further performance drops in real-world settings.

\subsubsection{Slow Reasoning for Anomaly Assessment \normalfont{(\textbf{H3})}}\label{sec:exps-slow-reasoning}
Once an \textit{anomaly} has been identified, it is the role of the slow reasoner to assess whether or not the anomaly warrants the enactment of safety-preserving interventions. 
Recall, we use LLMs for their generalist knowledge---acquired through internet-scale pretraining---to infer the need for a fallback on an observation identified as dissimilar from the robot's previous experiences.

\begin{table}[H]
    \centering
    \caption{
        Slow Generative Reasoning for Anomaly Assessment in VTOL. Best scores are bolded; second best are underlined.
    }
    \begin{tabular}{lccc}
        \toprule
        Method & TPR & FPR & Accuracy \\
        \midrule
        Llama 2 (7B)    & 0.52             & 0.46             & 0.52 \\
        GPT-3.5 Turbo & \textbf{0.97} & 0.54 & 0.73 \\
        GPT-3.5 Turbo CoT & 0.82 & 0.28 & 0.77 \\
        GPT-4       & 0.65    & \textbf{0.06} & \underline{0.79} \\
        GPT-4 CoT & \underline{0.89} & \underline{0.10} & \textbf{0.90} \\
        \bottomrule
    \end{tabular}
    \label{tab:synthetic-slow-reasoning}
\end{table}

\begin{figure*}[]
    \centering
    \includegraphics[width=0.92\textwidth]{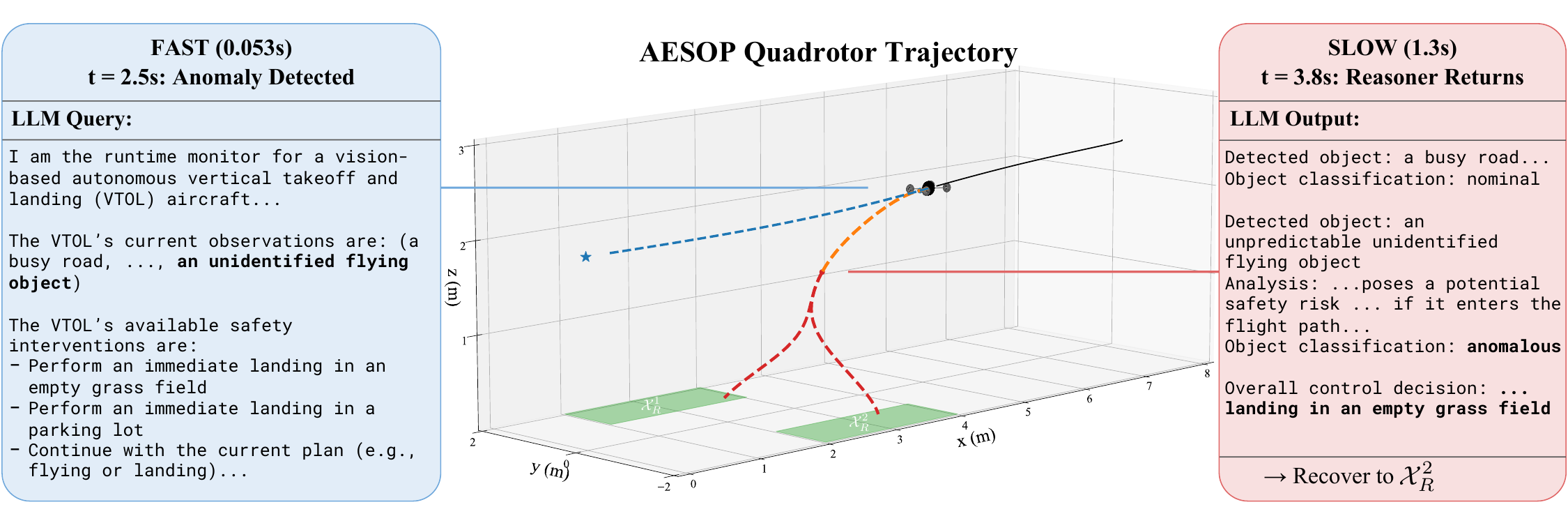}
    \caption{Closed-loop trajectory of a quadrotor using the AESOP algorithm. The figure represents a snapshot of the quadrotor at $t=2.5\mathrm{s}$: The trajectory until time $t$ is in black. The nominal trajectory plan is shown in blue, with a blue dot denoting the first consensus constraint in \cref{eq:modification-mpc}. The overlapping recovery trajectory plans, up to the consensus horizon corresponding to the LLM latency $K$, are in orange. The recovery trajectory plans deviate after $K$, shown in red, and they each reach their respective recovery region (in green). The blue text callout shows how the fast anomaly detector issues a warning and triggers the slow reasoner at $t=2.5\mathrm{s}$. The red callout shows the response from the slow reasoner, which the LLM returns within the $K$ consensus timesteps in the recovery plans.}
    \label{fig:aesop-drone}
\end{figure*}

We evaluate the ability of LLMs to assess anomalies in the VTOL domain and predict one of two options: whether the VTOL should 1) ``continue'' it's nominal operation or 2) ``fallback,'' enacting one of several listed safety-preserving interventions. 
To do so effectively, the LLM must differentiate between safety-redundant (e.g., a plane on a known flight path) and safety-concerning (e.g., a fighter jet) anomalies.
For this experiment, we evaluate four LLMs: Llama 2 (7B) single-token prediction, GPT-3.5 Turbo CoT, GPT-4 single-token prediction, and GPT-4 CoT.
As before, the CoT approach must first assess the safety risk each concept contained in the observation before outputting a fallback classification.

We report true positive rate (TPR), false positive rate (FPR), and accuracy.  
A true positive corresponds to the LLM correctly engaging a fallback intervention on a safety-concerning anomaly, while a true negative corresponds to correctly dismissing a safety-redundant anomaly.


The results are shown in \cref{tab:synthetic-slow-reasoning}.
We observe that in these out-of-distribution scenarios, model capacity is an important consideration, with the GPT variants clearly outperforming the Llama 2 (7B) baseline.
As expected, we also find that CoT reasoning yields a notable improvement in overall classification accuracy (e.g., $11$\% for GPT-4) and reduces the number of false positives.
These findings are corroborated in the manipulation domain (\cref{tab:synthetic-slow-reasoning-additional}).
This validates our third hypothesis \textbf{H3}.

\subsection{Full Stack---Quadrotor Simulation \normalfont{(\textbf{H4})}}\label{sec:exps-drone}
\begin{table}[b ]
\centering
\caption{Percentage of trajectories where the quadrotor successfully recovered to the LLM's choice of recovery region.}
\label{tab:drone-ablate}
\begin{tabular}{p{2cm}ccc}
\toprule
 & Naive MPC & FS-MPC \cite{SinhaSchmerlingEtAl2023} & AESOP \\ 
\midrule
Successful Recovery Rate &  15\% &  23\% & \textbf{100\%} \\
\bottomrule
\end{tabular}
\end{table}
Here, we demonstrate the efficacy of our framework, from anomaly detection and LLM reasoning to closed-loop control with Algorithm \ref{alg:safe-mpc}, on an example of a quadrotor delivering a package. In this simulation, the quadrotor's task is to fly toward and subsequently land at a target location. To simulate a variety of safety hazards and instantiate the fast anomaly detector, we recycle the VTOL synthetic observations and show the quadrotor anomalous observations at random time intervals. In this simulation, we encode the location of two landing regions, a parking lot and an open field, as polytopic state constraints representing $\calX_R^{1,2}$. To instantiate the MPC in \cref{eq:modification-mpc}, we use a fixed dynamics model discretized at a timestep of $\mathrm{dt}=.1\mathrm{s}$, with a planning horizon of $T = 4\mathrm{s}$. We assume that the slow reasoning LLM may take up to $K = 1.5\mathrm{s}$ to return an output, a number consistent with the latency of cloud querying GPT-4 reported in \cite{tagliabue2023real} using a conversational chat-based prompting approach. 
\begin{figure}[b]
    \centering
    \includegraphics[width=\linewidth]{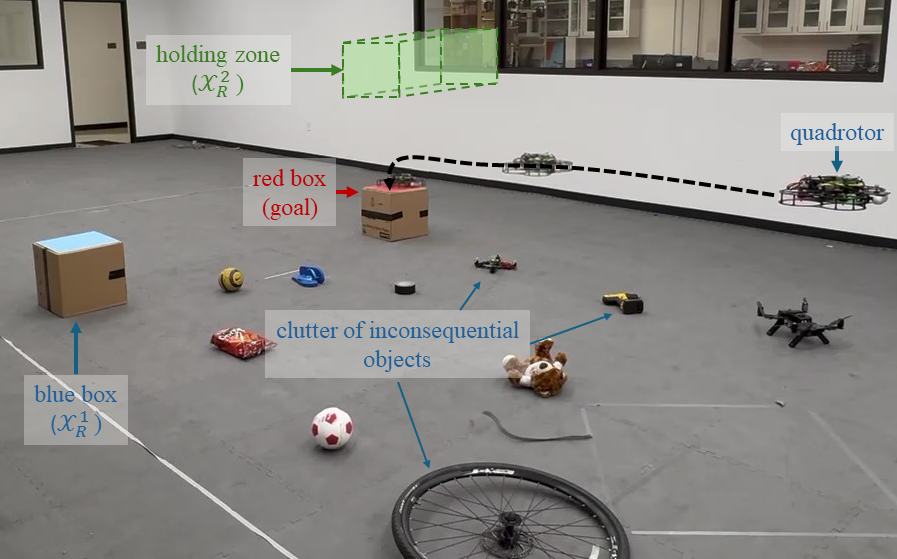}
    \caption{Annotated depiction of our quadrotor hardware experiment. The quadrotor's goal is to land on the red box. In the event of an anomaly, it can either recover by landing on the blue box, or by hovering within the designated holding zone.}
    \label{fig:hardware-description}
\end{figure}

In \cref{fig:aesop-drone}, we show a snapshot of the closed-loop trajectory of the quadrotor as it flies towards the landing zone. \cref{fig:aesop-drone} shows that the AESOP MPC \cref{eq:modification-mpc} plans two recovery trajectories that safely abort the control task and land the drone in their respective recovery regions, while still allowing the quadrotor to make progress towards its nominal objective. Furthermore, the recovery trajectories are aligned for the first $K$ timesteps, budgeting time for the LLM to output which, if any, of the recovery trajectories should be executed. The trajectories in \cref{fig:aesop-drone} show that the drone descends and slows down during the first $K$ timesteps of the recovery trajectories, thereby explicitly budgeting time for the LLM to reason.  


While we only show a single example in \cref{fig:aesop-drone} to illustrate the qualitative behavior of our method, we include additional plots in \cref{sec:drone} and videos of closed-loop trajectories on the \href{https://sites.google.com/view/aesop-llm}{project page} to more extensively demonstrate our approach. Furthermore, \cref{tab:drone-ablate} shows the results of a quantitative ablation over a set of $500$ scenarios and compares AESOP with 1) the fallback-planning method in \cite{SinhaSharmaEtAl2022} (which ignores the runtime monitor's latency), and 2) a naive planner that only tries to compute a recovery plan post-hoc after detecting a dangerous event. We refer to \cref{sec:drone} for a detailed description of these experiments and baselines. While the FSMPC algorithm mildly improves over the naive baseline, \cref{tab:drone-ablate} showcases the impact of accounting for the slow reasoner's latency within the control design. Moreover, throughout the closed-loop trajectory in \cref{fig:aesop-drone}, the average speed of the quadrotor is around $2.5 \mathrm{m/s}$, demonstrating that our framework allows for dynamic control of the robot while leveraging the slower LLM to improve safety in a reactive, real-time manner. This supports our fourth hypothesis \textbf{H4}.

\subsection{Quadrotor Hardware Demonstration \& Timing \normalfont{(\textbf{H4})}}\label{sec:exps-hardware}
Furthermore, we conduct hardware experiments with a physical quadrotor equipped with a downward facing camera (Intel Realsense D435). The quadrotor's nominal goal is to land on a designated red box amidst a cluttered environment. As shown in \cref{fig:hardware-description}, the ground is scattered with various objects, such as a bicycle tire, a soccer ball, and a drill, which are inconsequential to the landing task. \cref{fig:hardware-description} also shows the quadrotor's two recovery strategies to avoid failures when e.g., another quadrotor has landed on the red box: 1) landing at an alternative landing site and 2) to hover in a designated holding zone.

In order to construct the embedding cache for the fast detector, we first record images by flying the drone in a circular pattern above the operational area with a nominal clutter of objects on the ground. We use the open vocabulary object detector OWL-ViT\cite{minderer2205simple} to extract context-aware descriptions of visible objects (e.g.,``on the red box" or ``on the ground") from the image observations, which we then use to construct prompts for the embedding model (MPNet) and generative reasoner (GPT-3.5-Turbo). 


We evaluate our system on the following three scenarios. For more details on these scenarios, see \cref{app:hardware} and the videos on the \href{https://sites.google.com/view/aesop-llm}{project page}.

\textbf{1. Nominal Operation:} There are no obstructions on the red box, so the quadrotor lands normally despite the clutter.

\textbf{2. Consequential Anomaly:} We consider two variants of this scenario. In the first, another quadrotor has already landed on the red box, necessitating a diversion to the blue box for landing. In the second, other quadrotors occupy both red and blue boxes, necessitating a diversion to the holding zone. 

\textbf{3. Inconsequential Anomaly:} A previously unseen object (specifically, a keyboard) on the ground triggers the fast anomaly detector, after which the LLM correctly decides to proceed with landing at the nominal site.


\begin{table}[t]
\centering
\caption{Inference times for the OWL-ViT object detector, LLM embedding models (\texttt{MPNet}, \texttt{Mistral}), and cloud-querying \texttt{GPT-3/4} on a Jetson AGX Orin module.}
\label{my-table}
\begin{tabular}{lcc}
\toprule
Component & Mean (s) & Standard Deviation (s) \\ 
\midrule
MPC solve of \cref{eq:modification-mpc} & 0.023 & 0.019\\
\midrule
OWL-ViT & 0.025 & 0.002 \\
MPNet & 0.028 & 0.005 \\
Mistral & 0.32 & 0.08 \\
GPT-3-Turbo CoT & 3.10 & 0.85 \\
GPT-4 CoT & 18.88 & 3.923 \\
\bottomrule
\label{tab:jetson-timings}
\end{tabular}
\end{table}

In addition, we evaluate the computational cost of our pipeline on our hardware platform, an Nvidia Jetson AGX Orin, which is designed for embedded systems like a quadrotor. \cref{tab:jetson-timings} shows that the OWL-ViT detection parsing and MPNet embedding computation can jointly run at $18.8 \mathrm{Hz}$. This ensures that AESOP (Algorithm~\ref{alg:safe-mpc}) can comfortably operate at approximately $10\mathrm{Hz}$\footnote{In all our experiments, the computational cost of computing similarity scores with the embeddings was negligible compared to model inference times.}. We cloud query GPT-3.5-turbo with chain-of-thought prompting for the slow reasoning process, which, as shown in \cref{tab:jetson-timings}, has a non-negligible latency. Together with the simulations in \cref{sec:exps-drone}, these hardware results demonstrate that our approach effectively leverages LLMs to improve robot reliability despite their inference costs, thereby validating our hypothesis \normalfont{(\textbf{H4})}. 
We include further hardware results, detailing 1) an additional evaluation of the fast anomaly detector, 2) a analysis of LLM query latencies informing our choice of $K$, 3) implementation details of the MPC solver in \cref{app:hardware}.


\subsection{Ablation---Autonomous Vehicles Simulation}\label{sec:exps-ablation}
We run ablations on self-driving scenarios curated in \cite{elhafsi2023}. Here, CARLA was used to generate anomalous observations inspired by documented failure modes of self-driving perception systems\footnote{Examples of documented failures modes include an image of a stop sign on a billboard \hyperlink{https://www.youtube.com/watch?v=-OdOmU58zOw}{(source)} and a truck transporting inactive traffic lights \hyperlink{https://futurism.com/the-byte/tesla-autopilot-bamboozled-truck-traffic-lights}{(source)}.}.
All anomalies in this domain require safety-preserving intervention on the nominal system's operation.
Thus, this experiment resembles the synthetic fast reasoning experiments (\cref{sec:exps-fast-reasoning}), but additionally considers high-dimensional RGB observations as inputs to the runtime monitors.

\subsubsection{End-to-End Reasoning for Anomaly Detection \normalfont{(\textbf{H5})}}
Given the release of multi-modal models such as GPT-4V, we ablate performance differences between a two-step pipeline and a single-step pipeline. 
The two-step pipeline constructs a prompt consisting of detections from the open vocabulary OWL-ViT object detector~\cite{minderer2205simple} on CARLA images, whereas the single-step pipeline directly queries GPT-4V to output an anomaly classification based on the image observation.
For each of these approaches, we consider both single-token and CoT reasoning. 

The results are presented in \cref{tab:carla-anomaly-detection}.
In both text- and vision-based anomaly detection, we corroborate the notion that CoT reasoning facilitates more accurate responses to a complex task compared to single-token reasoning.
Furthermore, the results of GPT-4V (relative to GPT-3/4) suggest that the anomaly detection task could be performed end-to-end.

\begin{table}[t]
    \centering
    \caption{CARLA Evaluation. Text and Vision-based Anomaly Detection.}
    \label{tab:carla-anomaly-detection}
    \begin{tabular}{clccc}
        \toprule
        \textbf{} & Method & TPR & FPR & Bal. Accuracy \\
        \midrule
        \multirow{5}{*}{\rotatebox[origin=c]{90}{Text}} 
            & GPT-4 & 0.74 & 0.19 & 0.78\\
            & GPT-3 CoT~\cite{elhafsi2023} & 0.89 & 0.26 & 0.82\\
            & MPNet (Ours) & 0.69 & \textbf{0.05} & 0.82\\
            & Mistral (Ours) & 0.95 & \textbf{0.05} & \textbf{0.95}\\
        \midrule
        \multirow{6}{*}{\rotatebox[origin=c]{90}{Vision}} 
            & SCOD & 0.40 & 0.06 & 0.67\\
            & Mahal. & 0.40 & 0.13 & 0.64\\
            & GPT-4V & \underline{0.97} & 0.27 & 0.85\\
            & GPT-4V CoT & 0.89 & \underline{0.10} & \underline{0.90} \\
            & CLIP (Ours) & 0.86 & \textbf{0.05} & \underline{0.90} \\
            & CLIP (Ours) Abl. & \textbf{0.99} & 0.57 & 0.71\\
        \bottomrule
    \end{tabular}
\end{table}

\subsubsection{End-to-end Embedding-Based Anomaly Detection (\textbf{H5})}
For the following embedding evaluations, we use the top performing language embedding models, MPNet (110M) and Mistral (7B), operating on the existing prompts which list detected objects parsed from the output of OWL-ViT. Additionally, we evaluate image embeddings from a ViT, specifically CLIP~\cite{radford2021learning}, that provides direct visual grounding for the task. As in \cref{sec:exps-fast-reasoning}, we use top-$5$ scoring across all methods.


To ensure we fairly compare the expressiveness of purely language-based embeddings with vision-language embeddings, we report results using ground-truth object detections to construct the prompts for the text-based embedding models in \cref{tab:carla-anomaly-detection}, in the spirit of a more comprehensively engineered system with reliable object detection. We do so because, as noted in \cite{elhafsi2023}, the vision model suffers from a real-to-sim domain shift, and sometimes tends to, e.g., characterize nominal observations of stop signs as ``images of stop signs''. We include an ablation showing the impact of vision errors in \cref{sec:ablation}.

First, we observe that language embeddings from the relatively small MPNet model are less capable at discerning anomalies among many nominal observations than Mistral, independent of whether OWL-ViT returns a correct scene description (see \cref{sec:ablation}). 

Second, we surpass GPT-4 and GPT-4V with Mistral (roughly $64$x larger than MPNet).
Besides suggesting that a smaller model's language embeddings are less semantically rich and therefore less capable at capturing the presence of more subtle anomalies, we also show in \cref{sec:ablation} that MPNet's accuracy degrades as the number of objects in an observation increases, whereas Mistral's accuracy remains consistent when the number of detections in an image varies.

Third, we note that the use of CLIP embeddings derived directly from the vehicle's RGB observations also achieves high performance, comparable to GPT-4V CoT. 
This suggests that compute-intensive, CoT reasoning is not necessary to discern the anomalies directly from vision in semantically rich environments. However, we nuance this finding by noting that the observations in the CARLA dataset were constructed by driving a car along simulated routes and placing object assets that trick the vehicle into making unsafe decision along those routes \cite{elhafsi2023}. This means that the routes appear both with and without anomalous objects and that episodes wherein the vehicle takes unsafe actions include both nominal and anomalous observations. When we change the calibration strategy to only construct the embedding cache with nominal observations from routes that never pass by anomalous objects (denoted by ``Abl.'' in \cref{tab:carla-anomaly-detection}), we find that CLIP's FPR increases significantly. This is because CLIP embeddings contain a mix of visual and semantic features \cite{KaramchetiEtAl2023}, and therefore the slight visual novelties in an unseen route are flagged as anomalous even though they are semantically uninteresting. As such, CLIP's limitations are related to the SCOD \cite{sharma2022} and Mahalanobis distance \cite{LeeEtAl2018} baseline OOD detectors (taken from \cite{elhafsi2023}) that wrap the AV's base object detector (DETR \cite{CarionEtAl2020}). In contrast, as we further ablate in \cref{sec:ablation}, the two-stage detection approach is unaffected by differences in visual appearance. 
Despite this, and noting that further work should examine how to disentangle semantic and visual features to increase robustness, we argue that our preliminary findings on using multi-modal embeddings directly offers significant promise for streamlining the implementation of our framework.
This validates our fifth and final hypothesis \textbf{H5}.


\section{Conclusion and Outlook}\label{sec:outlook}
In this paper, we presented a runtime monitoring framework utilizing generalist foundation models to facilitate safe and real-time control of agile robotic systems faced with real-world anomalies.
This is enabled through a reasoning hierarchy: a fast anomaly classifier querying similarity with the robot's prior experiences in an LLM embedding space, and a slow generative reasoner assessing the safety implications of detected anomalies and selecting the appropriate mitigation strategy.
These reasoners are interfaced with a new model predictive control strategy that maintains the feasibility of multiple safe recovery plans.  

In extensive experiments, we demonstrate that a) embedding-based anomaly detection performs favorably to zero-shot generative reasoning with high-capacity LLMs, thanks in part to the grounding afforded by the prior embedding experience of the robot; b) embedding-based anomaly detection attains strong performance even when instantiated with small language models, allowing our method to run onboard computationally constrained robotic systems; c) dual-stage reasoning enables LLMs to operate in the real-time reactive control loop of an agile robot; d) alternative forms of embeddings, such as those obtained from vision-based foundation models, can be used to efficiently detect anomalies in high-dimensional observation spaces.

As such, our work highlights the potential of LLMs and, more broadly, foundation models toward significant increases in the robustness of autonomous robots with respect to unpredictable and unusual out-of-distribution scenarios or \textit{tail events}.
Improving the performance and generality of our framework presents several promising avenues for future research.
For example, the impact of LLM inference latencies could be reduced by devising methods to constrain generative reasoning to a fixed word budget, or by using intermediate generations to inform decision-making during the generation process.
Further analysis is required on the \textit{correctness} of fallback plans selected by the LLM, and whether fallbacks can be programmatically determined upon latency timeout. Finally, continual learning based on the delayed anomaly assessment of the generative reasoner could be used to avoid triggering the slow reasoner on non-safety-critical anomalies a second time.

\section*{Acknowledgments}
The authors would like to thank Brian Ichter and Fei Xia for insightful discussions and feedback throughout the project. In addition, the authors are indebted to Jun En Low, Keiko Nagami, and Alvin Sun for their assistance in setting up the hardware experiments. The NASA University Leadership initiative (grant \#80NSSC20M0163) and the Toyota Research Institute (TRI) provided funds to assist the authors with their research, but this article solely reflects the opinions and conclusions of its authors and not any NASA or TRI entity.


\bibliographystyle{plainnat}
\bibliography{references}

\include{supp.tex}

\end{document}

%% file: commands.tex
\input{calfabet}

\newcommand{\todo}[1]{{\color{red}{TODO: #1}}}
\newcommand{\rohan}[1]{{\color{orange}{Rohan: #1}}}
\newcommand{\chris}[1]{{\color{orange}{Chris: #1}}}
\newcommand{\amine}[1]{{\color{purple}{Amine: #1}}}
\newcommand{\matt}[1]{{\color{olive}{Matt: #1}}}

\newcommand{\E}{\mathbb{E}}
\newcommand{\p}{\mathbb{P}}
\newcommand{\R}{\mathbb{R}}
\newcommand{\Ps}{\mathcal{P}}

\newcommand{\calhA}{\widehat{\mathcal{A}}}
\newcommand{\calhU}{\widehat{\mathcal{U}}}

\newcommand{\Qhat}{\widehat{Q}}
\newcommand{\Vhat}{\widehat{{V}}}

\newcommand{\prob}{\mathbb{P}}
\newcommand{\iid}{\overset{\textup{iid}}{\sim}}

\newcommand{\x}{\bm{x}}
\newcommand{\y}{\bm{y}}
\newcommand{\z}{\bm{z}}
\newcommand{\f}{\bm{f}}
\newcommand{\g}{\bm{g}}
\newcommand{\h}{\bm{h}}
\newcommand{\e}{\bm{e}}
\newcommand{\w}{\bm{w}}
\newcommand{\s}{\bm{s}}
\newcommand{\obs}{\bm{o}}
\newcommand{\vn}{\bm{v}}
\newcommand{\fu}{\bm{u}}
\newcommand{\traj}{\bm{\tau}}
\newcommand{\params}{\bm{\theta}}

\newcommand{\pre}{\mathrm{Pre}}

\newcommand{\nn}{\mathrm{NN}}

\newtheorem{claim}{Claim}
\newtheorem{remark}{Remark}
\newtheorem{lemma}{Lemma}
\newtheorem{theorem}{Theorem}
\newtheorem{corollary}{Corollary}
\newtheorem{definition}{Definition}
\newtheorem{assumption}{Assumption}
\newtheorem{example}{Example}

\newcommand{\comment}[1]{{\begin{quotation}\color{red}``{#1}''\end{quotation}}}
\newcommand{\inlinecomment}[1]{{\color{red}``{#1}''}}
\newcommand{\edit}[1]{{\begin{quotation}\color{blue}{#1} \end{quotation}}}
\newcommand{\inlineedit}[1]{{#1}}
\newcommand{\original}[1]{{\begin{quotation}\color{magenta}{#1}\end{quotation}}}

%% file: calfabet.tex
\newcommand{\calZ}{\mathcal{Z}}
\newcommand{\calY}{\mathcal{Y}}
\newcommand{\calX}{\mathcal{X}}
\newcommand{\calW}{\mathcal{W}}
\newcommand{\calV}{\mathcal{V}}
\newcommand{\calU}{\mathcal{U}}
\newcommand{\calT}{\mathcal{T}}
\newcommand{\calS}{\mathcal{S}}
\newcommand{\calR}{\mathcal{R}}
\newcommand{\calQ}{\mathcal{Q}}
\newcommand{\calP}{\mathcal{P}}
\newcommand{\calO}{\mathcal{O}}
\newcommand{\calN}{\mathcal{N}}
\newcommand{\calM}{\mathcal{M}}
\newcommand{\calL}{\mathcal{L}}
\newcommand{\calK}{\mathcal{K}}
\newcommand{\calJ}{\mathcal{J}}
\newcommand{\calI}{\mathcal{I}}
\newcommand{\calH}{\mathcal{H}}
\newcommand{\calG}{\mathcal{G}}
\newcommand{\calF}{\mathcal{F}}
\newcommand{\calE}{\mathcal{E}}
\newcommand{\calD}{\mathcal{D}}
\newcommand{\calC}{\mathcal{C}}
\newcommand{\calB}{\mathcal{B}}
\newcommand{\calA}{\mathcal{A}}

%% file: supp.tex
\appendix

These appendices contain further details on the experiments in the main body of the paper, additional results and ablations supporting the main hypotheses of the paper, analysis of these supplementary results, and proofs of theoretical results. Besides the results within this document, we also refer the reader to 
\textbf{videos of the quadrotor experiments in the \texttt{quad\_videos/}} 
directory of the supplemental material. 
Furthermore, we include videos describing our approach, prompt templates, and further results on our \textbf{project page: \url{https://sites.google.com/view/aesop-llm}}. 
These appendices are organized as follows:

\localtableofcontents

\inlineedit{
\subsection{Notation and Glossary}\label{ap:glos}

We include a glossary of all the notation and symbols used in this paper in Table \ref{tab:glos}.

\begin{table}[!h]
    \centering

    \inlineedit{
    \begin{tabular}{p{.15\linewidth}p{.15\linewidth}p{.6\linewidth
    }}
    \hline
    & Symbol & Description \\
\hline
   & $x$ & unless explicitly defined otherwise, scalar variables are lowercase \\
   & $\mathbf{x}$ & vectors are boldfaced \\
   & $\calX$ & sets are caligraphic \\ 
   & $x_t$ & time-varying quantities are indexed with a subscript $t \in \mathbb{N}_{\geq 0}$\\ 
 Notation and conventions & $\x_{0:t}$ & Shorthand to index subsequences: $\x_{0:t} := \{\x_0,\dots, \x_t\}$ \\
   & $\lambda, \delta, \epsilon, \theta$ & hyperparameters (regardless of their type) are lowercase Greek characters \\
   & $\x_{t+k|t}$ & Predicted quantities at $k$ time steps into the future computed at time step $t$. Read $\x_{t+k|t}$ as ``the predicted value of $\x$ at time $t + k$ given time $t$.'' \\
    \hline
    \hline
    & $\x$ & System state \\
    & $\fu$ & Input \\
    & $\obs$ & Observation \\
    & $\f$ & Dynamics \\
    & $\h$ & Anomaly Detector \\
    & $\w$ & Generative reasoner \\
    &$\e$ & Embedding vector \\
    & $\phi$ & Embedding model \\
    & $\calX$ & State constraint set \\
    & $\calU$ & Input constraint set \\
    & $\calO$ & Observation space \\
    & $\tau$ & Anomaly detection threshold \\
    & $\s$ & Anomaly score function \\
    & $C$ & control objective function for the MPC \cref{eq:modification-mpc}\\
    & $\calD_{\mathrm{nom}}$ & Dataset of nominal observations \\
    & $N$ & Number of nominal observations in $\calD_{\mathrm{nom}}$ \\
    Variables & $\calX_R^i$ & the $i$'th recovery region \\
    & $d$ & Number of recovery regions \\
    & $\calD_e$ & Embedding vector cache constructed from $\calD_{\mathrm{nom}}$ \\
    & $\alpha$ & Quantile hyperparameter to select the anomaly detector threshold \\
    & $q$ & Optimization variable used to define the empirical $\alpha$-quantile \\
    & $\calY$ & Subset of $\{1, \dots, d\}$ indicating a selection of recovery regions \\
    & $K$ & Upper bound on the latency of the slow reasoner \\
    & $T$ & Time horizon of the MPC \cref{eq:modification-mpc} \\
    & $J$ & Objective value associated with the solution of the MPC problem \cref{eq:modification-mpc} \\
    & $\x^\star, \ \fu^\star$ & Starred quantities denote the optimal values of the decision variables in the MPC problem \cref{eq:modification-mpc} \\
    & $t_{\mathrm{anom}}$ & Time step at which the anomaly detector triggers \\
    \hline
     \end{tabular}
    \caption{Glossary of notation and symbols used in this paper.}
    \label{tab:glos}}
\end{table}

\subsection{Background: Anomaly Detection}\label{sec:prelim-anomaly}
In essence, an anomaly detector is a classifier $\h: \calO \to \{\mathrm{nominal}, \ \mathrm{anomaly}\}$ that maps observations to a detection at runtime. However, limited access to anomalous examples (after all, it is their dissimilarity from prior experiences that makes them anomalous) typically precludes us from training a classifier with an obvious decision boundary using supervised learning \cite{AngelopoulosBates2022, RuffKauffmanEtAl2021}. Instead, anomaly detection algorithms require two steps: First, we must construct a scalar score function $\s(o) \in \R$ from an observation, where a higher score indicates that the sample is ``more'' anomalous. Second, we need to calibrate a decision threshold $\tau \in \R$ on the score function such that: 
\begin{equation*}\label{eq:anomaly-classifier}
    \h(o) = \begin{cases} \mathrm{anomaly} &\text{if $\s(o) > \tau$} \\ \mathrm{nominal} &\text{if $\s(o) \leq \tau$} \end{cases}. 
\end{equation*} 
We investigate several score functions in this paper and compare their downstream utility in improving the safety and reliability of an autonomous robot. This necessitates instantiating the anomaly detectors with specific thresholds and measuring the accuracy of the subsequent calibrated classifier, since generic measures of a score functions' expressiveness are not guaranteed to capture the overall impact on an autonomy stack.}

\subsection{Background: Mechanics of LLMs}\label{sec:prelim-llms}
The decoder-only LLM architecture typically stacks together large numbers of Transformer modules to construct a mapping from a sequence of input tokens $x_{0:t}$ to a contextual embedding matrix $\phi(x_{0:t}) \in \R^{n \times t+1}$. That is, each input token gets mapped to a corresponding contextual embedding. It is well-known that the contextual embeddings are generally useful for prediction tasks themselves and often exhibit interesting properties. For example, some models are trained with contrastive losses to ensure embeddings with similar semantic meaning cluster closely in the embedding space, enabling retrieval of relevant information via similarity search. However, current research in robotics focuses on using LLMs to generate strings of text through autoregressive generation, i.e., next token prediction. To do so, a linear classification head is typically added onto the output contextual embedding to define a probability distribution over the next token
\begin{equation}
    x_{t+1} \sim p_{\mathrm{llm}}(x_{t+1} | \phi(x_{0:t})) := \mathrm{softmax}(W_{\mathrm{out}} \phi(x_{0:t})_t),
\end{equation}
which is then sampled and appended to the token sequence, after which the next token can be sampled. It should be clear that generating output sequences carries a computational cost that scales superlinearly in the cost of computing an embedding. That is, the zero-shot reasoning capabilities of LLM generation come at a computational cost, whereas direct learning on embeddings requires a source of supervision.

\subsection{Additional Results: Synthetics}\label{supp:synthetics}
The main goal of the \textit{synthetics} experiments is to analyze the performance characteristics of our fast anomaly detector across robotics domains with diverse observational and task semantics.
Here, we extend our synthetics results and analysis (\cref{sec:exps-synthetics}) to evaluate three such performance characteristics. 
First is the embedding-based anomaly detector's quality as measured by area under the receiver operating characteristic (AUROC) curve (\cref{supp:quality-analysis}). 
Second is the anomaly detector's sensitivity with respect to varying detection thresholds (\cref{supp:threshold-analysis}).
Lastly, we evaluate the anomaly detector's performance with respect to the choice of similarity score function (\cref{supp:score-fn-analysis}).

\begin{figure*}[tp]
    \centering
    \includegraphics[width=0.95\textwidth]{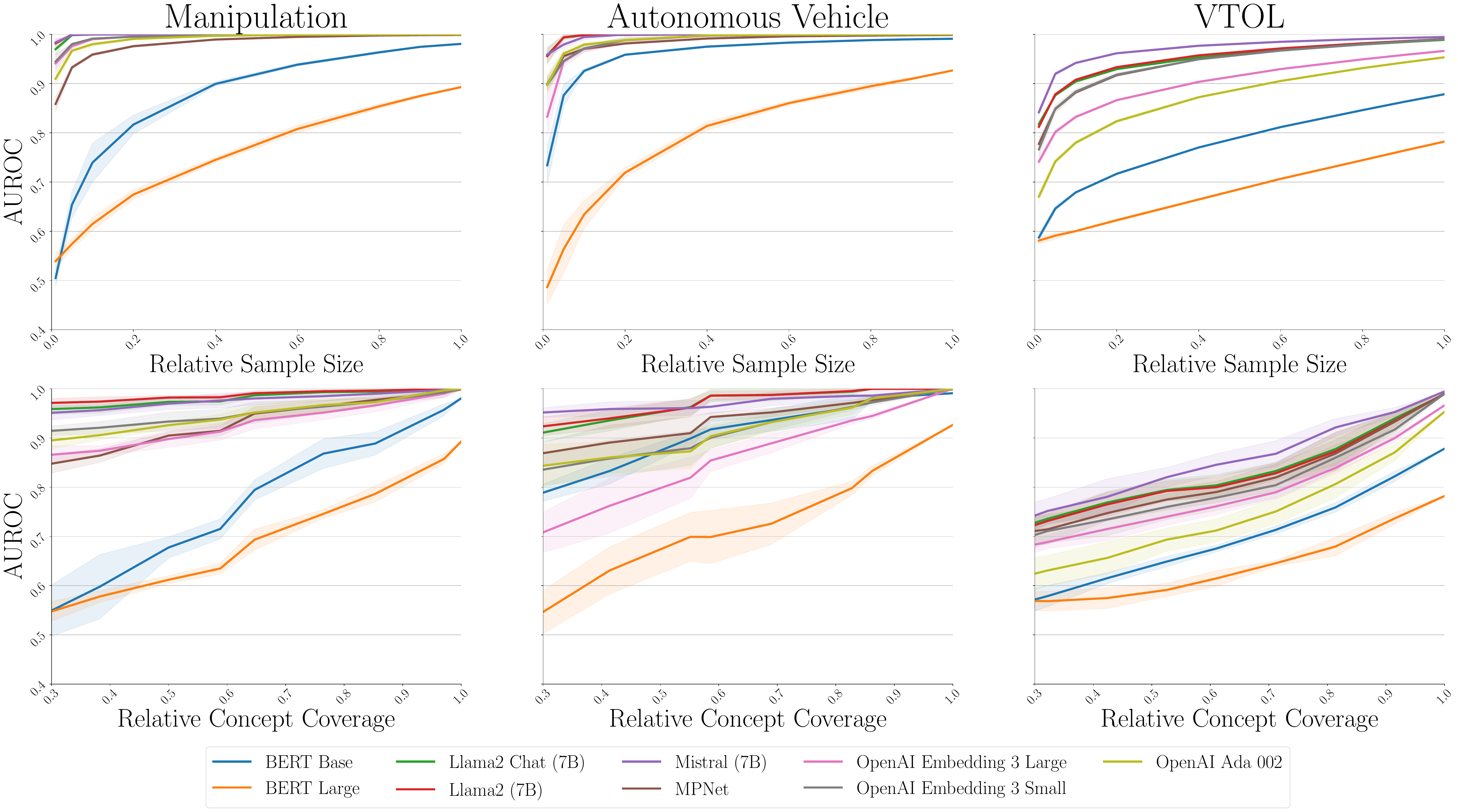}\\
    \caption{Embedding-based anomaly detection results for the manipulation, autonomous vehicle, and VTOL domains. The top row of figures plot the AUROC as a function of experiences sampled IID from the respective domain datasets. The bottom row of figures plot accuracy as a function of the concepts sampled from the respective domain datasets.}
    \label{fig:synthetic-roc-results}
\end{figure*}

\subsubsection{Fast Anomaly Detector ROC Analysis}\label{supp:quality-analysis}
We follow the synthetics evaluation scheme presented in the main results, which compares the performance of our fast anomaly detector over nine language models, using top-$5$ scoring, with a detection threshold set to the $95$-th quantile of the scores in the nominal dataset (\cref{eq:emp-quantile}).
Instead of reporting accuracy, we now measure performance in terms of AUROC, which more holistically reflects the detector's performance across varying detection thresholds.

The results are shown in \cref{fig:synthetic-roc-results}.
We observe similar performance trends to the previous results (\cref{fig:synthetic-results}, measured in terms of accuracy), where MPNet closely rivals the top performing 7B parameter models, Mistral and Llama 2, followed by the OpenAI embedding models, and lastly, the BERT models. 
These trends become increasingly clear as domain complexity increases (i.e., from the Manipulation to VTOL domains, left to right).
In the most challenging VTOL domain, we observe that concept coverage is key to attaining strong performance. 
This is perhaps a byproduct of the more nuanced concept shifts in complex domains, which necessitate comprehensive coverage of nominal concepts to deduce anomalies.
Consider how, in the VTOL domain, the anomalous ``swarming flock of birds'' can be mistakenly interpreted as \textit{similar} to the nominal ``flying bird'' without the additional grounding from other nominal concepts such as a ``blimp'' or ``quadcopter.''

\subsubsection{Detection Threshold Analysis}\label{supp:threshold-analysis}
To construct the anomaly detector (\cref{eq:anomaly-classifier}), we need to select or calibrate a detection threshold $\tau$ to differentiate anomalous from nominal observations.
Many techniques exist for calibrating detection thresholds, several of which offer specific guarantees on e.g., false positive rates, such as conformal prediction~\cite{AngelopoulosBates2022}.

In this experiment, we show the performance variation of our fast anomaly detector with respect to a range of detection thresholds (i.e., empirical quantiles) in an attempt to capture the \textit{sensitivity} of our method to, for example, well or poorly calibrated thresholds.
We report anomaly detection accuracy on the VTOL domain because, as shown by VTOL's relatively slowly increasing AUROC trends in \cref{fig:synthetic-roc-results}, we may expect larger variances in performance across thresholds.
We evaluate three language models, including OpenAI Ada 002, MPNet, and Mistral (7B), and randomly (IID) sample $80$\% of the full nominal dataset to construct the embedding cache for the anomaly detector.
Results are reported over $5$ random seeds.

The results are shown in \cref{tab:quantiles-results}. 
First, we observe a positive relationship between anomaly detection accuracy and threshold quantile across all methods. 
Once again, MPNet performs nearly identically to Mistral (7B), while OpenAI Ada 002 begins to plateau at the $85$-th quantile. 
Both MPNet and Mistral (7B) consistently outperform generative reasoning with GPT-4 (with the exception of the $75$-th quantile).
From the relatively small (yet non-negligible) performance improvements among increasing quantiles, we conclude that 1) while our anomaly detector is reasonably robust to the choice of detection threshold, 2) performance can be improved through the use of more sophisticated calibration techniques. 
Lastly, all methods produce negligible standard deviations across the random seeds, likely because the sampled embedding cache ($80$\% of the full nominal dataset) provides $\sim100$\% coverage over all nominal \textit{concepts}, which \cref{fig:synthetic-roc-results} shows heavily influences performance.

\begin{table*}[t]
\centering
\caption{Calibration results for a selection of embedding models in the VTOL domain. The table reports the mean anomaly detection accuracy thresholding the top-5 score function at different quantile thresholds. Standard deviations are provided in parentheses. Statistics were computed over multiple samplings of 80\% of the available nominal data samples. Accuracies for GPT-4 single-token and CoT queries are provided for comparison.}
\label{my-table}
    \begin{tabular}{lccccccc}
    \toprule
    & \multicolumn{4}{c}{\textbf{Embedding}} & & \multicolumn{2}{c}{\textbf{Generative}} \\
    \cmidrule{2-5} \cmidrule{7-8}
    Threshold: & 75-Quantile & 85-Quantile & 90-Quantile & 95-Quantile & & GPT-4 & GPT-4 CoT \\ 
    \midrule
    OpenAI Ada 002 & 0.8 (0.002) & 0.84 (0.002) & 0.85 (0.002) & 0.85 (0.002) & & \multirow{3}{*}{0.75} & \multirow{3}{*}{0.86} \\
    MPNet & \textbf{0.83} (0.002) & \textbf{0.89} (0.002) & \textbf{0.91} (0.002) & \underline{0.93} (0.001) & & & \\
    Mistral (7B) & \textbf{0.83} (0.001) & \textbf{0.89} (0.001) & \textbf{0.91} (0.001) & \textbf{0.94} (0.001) & & & \\
    \bottomrule
\label{tab:quantiles-results}
\end{tabular}
\end{table*}

\begin{figure*}[tp]
    \centering
    \includegraphics[width=0.95\textwidth]{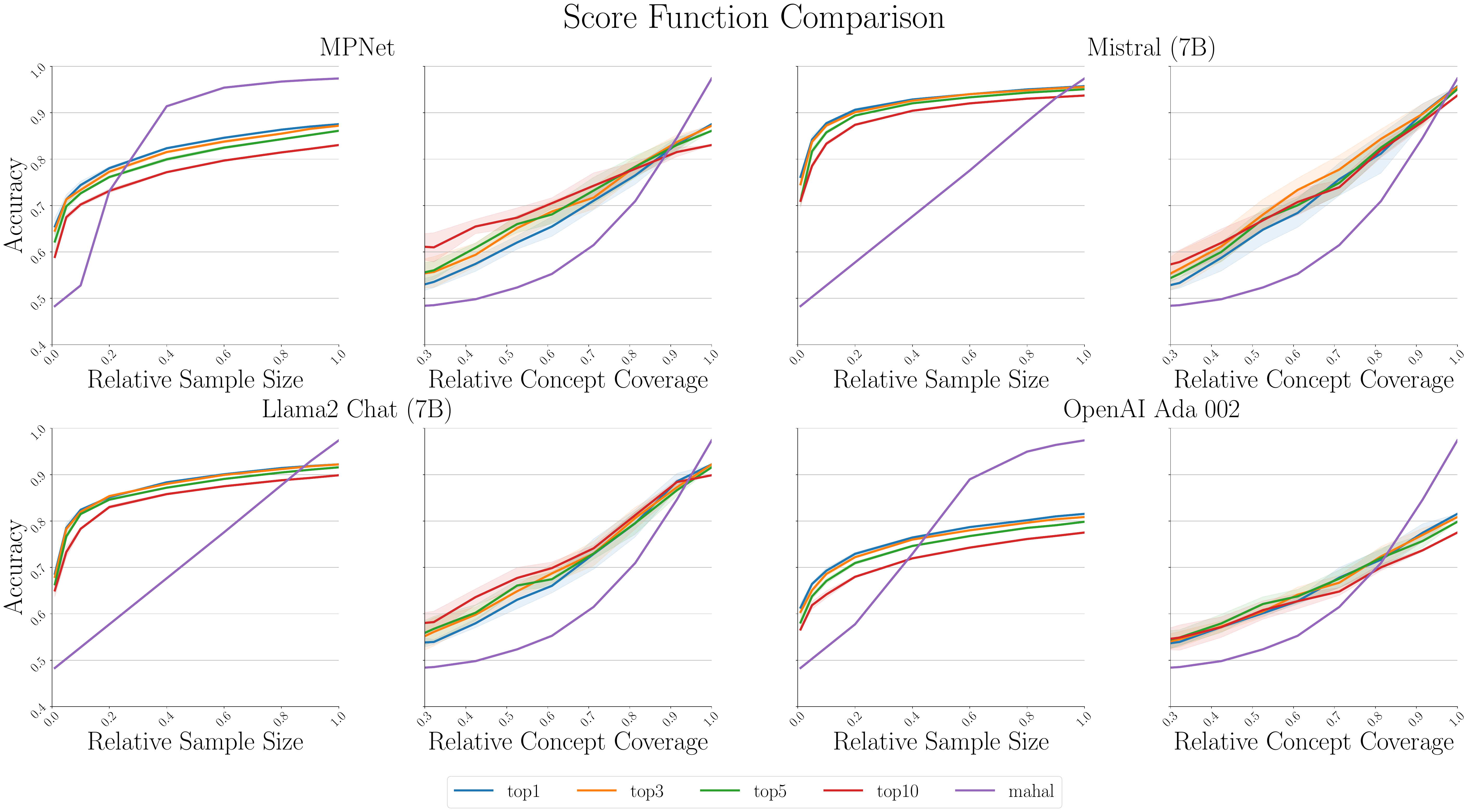}\\
    \caption{Score function comparison for a selection of embedding models applied to the VTOL domain. We compare the top-$k$ (for $k \in \{1,3,5,10\}$) and Mahalanobis distance score functions for MPNet, Mistral (7B), Llama2 Chat (7B), and OpenAI Ada 002.}
    \label{fig:score-fn-results}
\end{figure*}

\subsubsection{Similarity Score Function Analysis}\label{supp:score-fn-analysis}
Our fast anomaly detector can be instantiated with an arbitrary choice of similarity score function $\s(\e_t; \calD_e)$.
For brevity, our main results exclusively featured the use of top-$5$ scoring.
Thus, we conduct an ablation experiment to test the robustness of our approach to the choice of score function.

Our ablation includes top-$k$ scoring for $k$ in $\{1, 3, 5, 10\}$, which quantifies the distance between the input embedding $\e_t$ and the $k$ closest embeddings in the nominal embedding cache $\calD_e = \{\e_i\}_{i=1}^N$, and 2) the Mahalanobis distance score between the current embedding $\e_t$ and the multivariate Gaussian distribution formed from the mean and covariance of $\calD_e = \{\e_i\}_{i=1}^N$.
As before, we experiment on VTOL synthetic due to its size and complexity, evaluating four language models of varying size and function: OpenAI Ada 002, MPNet, Llama 2 Chat (7B), Mistral (7B). 

The results are shown in \cref{fig:score-fn-results}.
We first observe that the accuracy difference between all score functions at $100$\% concept coverage is within approximately $5$\% across all evaluated language models except OpenAI Ada 002. 
At first glance, this might suggest that the choice of score function is, to an extent, irrelevant for achieving high detection accuracies.
Upon closer analysis, we see that the top-$1$ and Mahalanobis score functions perform quite poorly in the low-data regime for most language models, while MPNet is able to utilize Mahalanobis better than others.
Overall, we find that top-$k$ for $k$ in $\{3, 5, 10\}$ are all strong choices of score functions that demonstrate competitive performance in several data regimes. 

\inlineedit{\subsubsection{Safety Assessment in Manipulation Domain}\label{appx:anomaly-assessment}

Recall, once an anomaly has been detected by the fast-reasoner, the slower reasoner is tasked with assessing whether the observation requires a safety-preserving fallback. In the main body of the report, we present results for a safety assessment of anomalies in VTOL domain with various SoTA language models. In \cref{tab:synthetic-slow-reasoning-additional}, we extend our safety assessment to the manipulation domain. Similar to the trend observed in \cref{fig:synthetic-results}, safety assessment is significantly easier in the manipulation domain than in the VTOL. Further, we see strong performance gains in using GPT-4 to correctly recognize an inconsequential anomaly in comparison to GPT-3.5.} 

\begin{table}[h]
    \centering
    \caption{\inlineedit{Slow Generative Reasoning for Anomaly Assessment in the Warehouse Manipulation Domain.}}
    \inlineedit{
    \begin{tabular}{clccc}
        \toprule
        Domain & Method & TPR & FPR & Accuracy \\
        \midrule
        \multirow{4}{*}{\rotatebox[origin=c]{90}{Manip.}}
            & GPT-3.5 Turbo     & \textbf{1.0} & 0.73 & 0.64 \\
            & GPT-3.5 Turbo CoT & \textbf{1.0} & 0.52 & 0.74 \\
            & GPT-4             & \textbf{1.0}  & \textbf{0.0} & \textbf{1.0} \\
            & GPT-4 CoT         & \textbf{1.0}  & \textbf{0.0} & \textbf{1.0} \\
        \bottomrule
    \end{tabular}}
    \label{tab:synthetic-slow-reasoning-additional}
\end{table}


        

\subsection{Additional Results: Quadrotor Simulation}\label{sec:drone}
In this section, we include additional details about our quadrotor simulation and describe our implementation of the MPC solver used in Algorithm \ref{alg:safe-mpc} (AESOP). We show the qualitative behavior and improvement of our algorithm in comparison with two baseline methods in \cref{fig:aesop-compare} in addition to \textbf{annotated videos of the trajectories in the supplementary files}. We also quantitatively evaluate the rate at which AESOP and the baselines successfully recover in over a set of $500$ randomized scenarios. 
\subsubsection{Simulation and Implementation Details} 
\begin{figure*}[p!]
    \begin{subfigure}[t]{0.47\textwidth}
    \includegraphics[width=\textwidth]{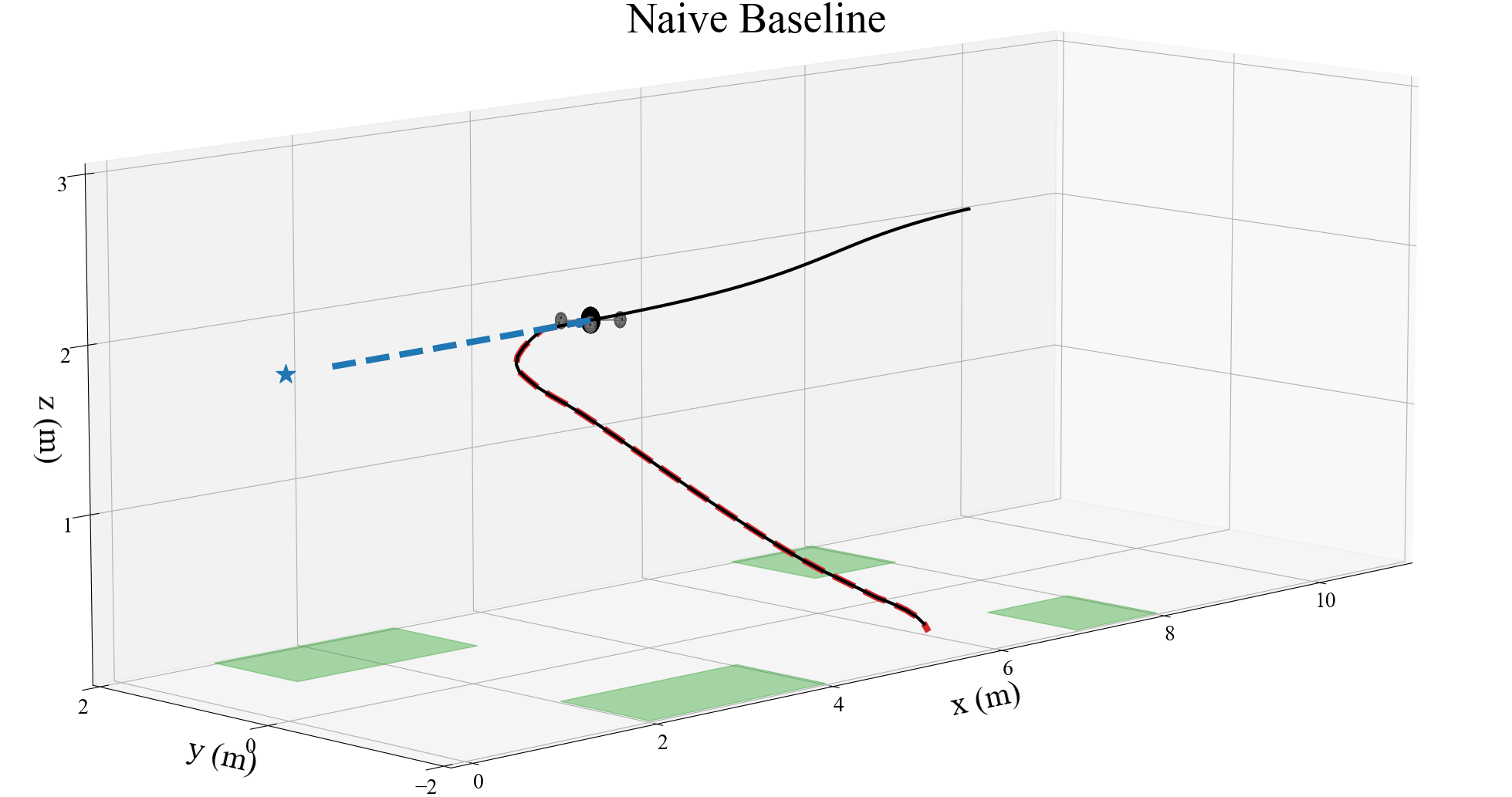}
        \caption{Closed loop trajectory of the naive baseline MPC (black). Also shown are the nominal predicted trajectory (blue) and the minimum constraint violating recovery trajectory (red) at the timestep that the slow LLM reasoner outputs.}
        \label{fig:naive-mpc}
    \end{subfigure}
    \hfill
    \begin{subfigure}[t]{0.47\textwidth}
    \includegraphics[width=\textwidth]{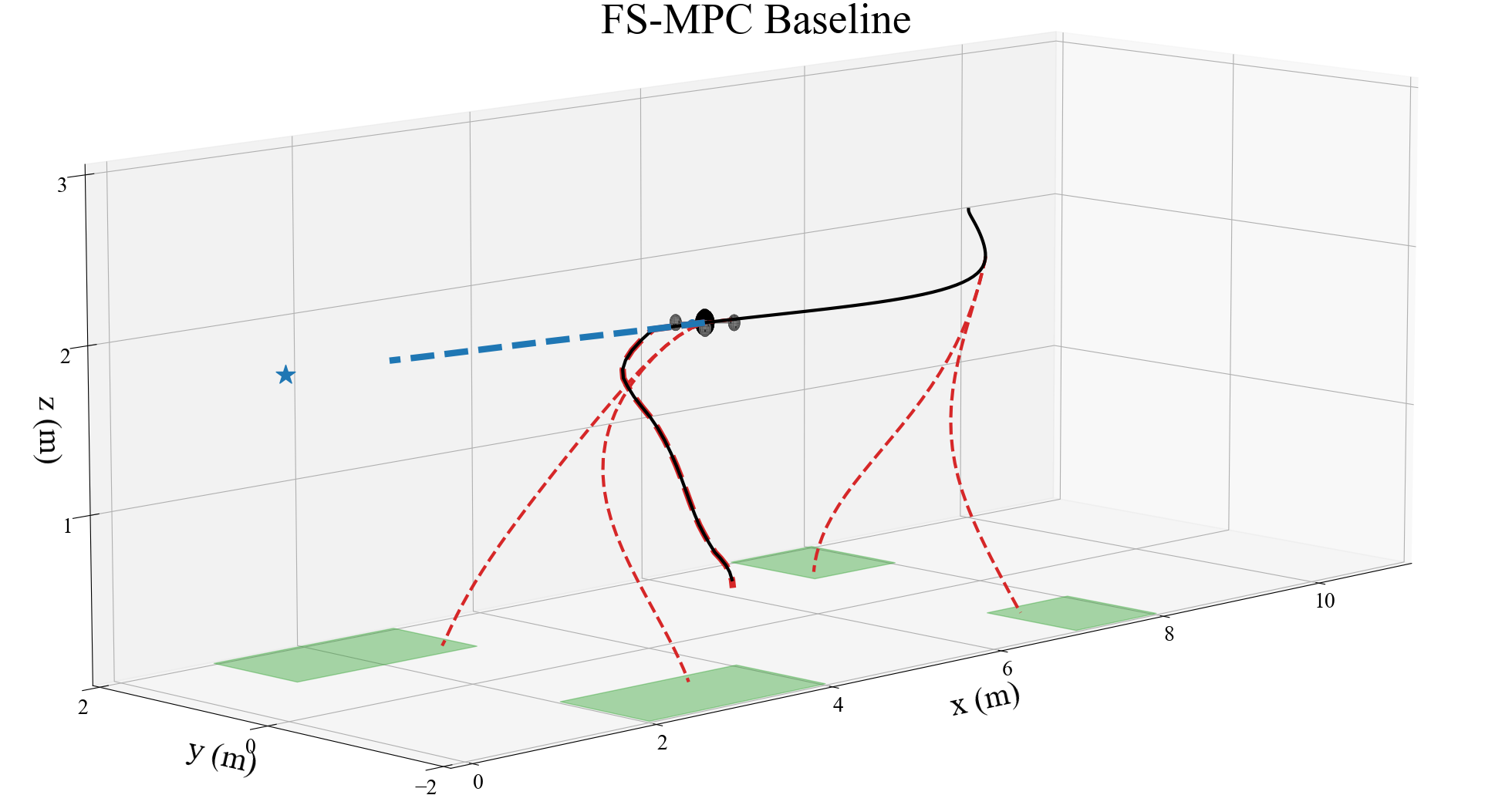}
        \caption{Closed loop trajectory of the Fallback-Safe MPC \cite{SinhaSchmerlingEtAl2023} (black). We show three recovery trajectory plans: The first is at the timestep where the LLM is queried, where the robot maintains recovery plans to the sets at $x\approx8\mathrm{m}$. The second set of plans is at the timestep right before the LLM reasoner outputs it's recovery decision, where the planner chooses recovery sets around $x\approx3\mathrm{m}$. The third is the minimum constraint violating recovery trajectory at the timestep that the slow LLM reasoner outputs.}
        \label{fig:fsmpc}
    \end{subfigure}
    \vspace{15pt}
    \\
     \centering
    \begin{subfigure}[t]{0.47\textwidth}
    \includegraphics[width=\textwidth]{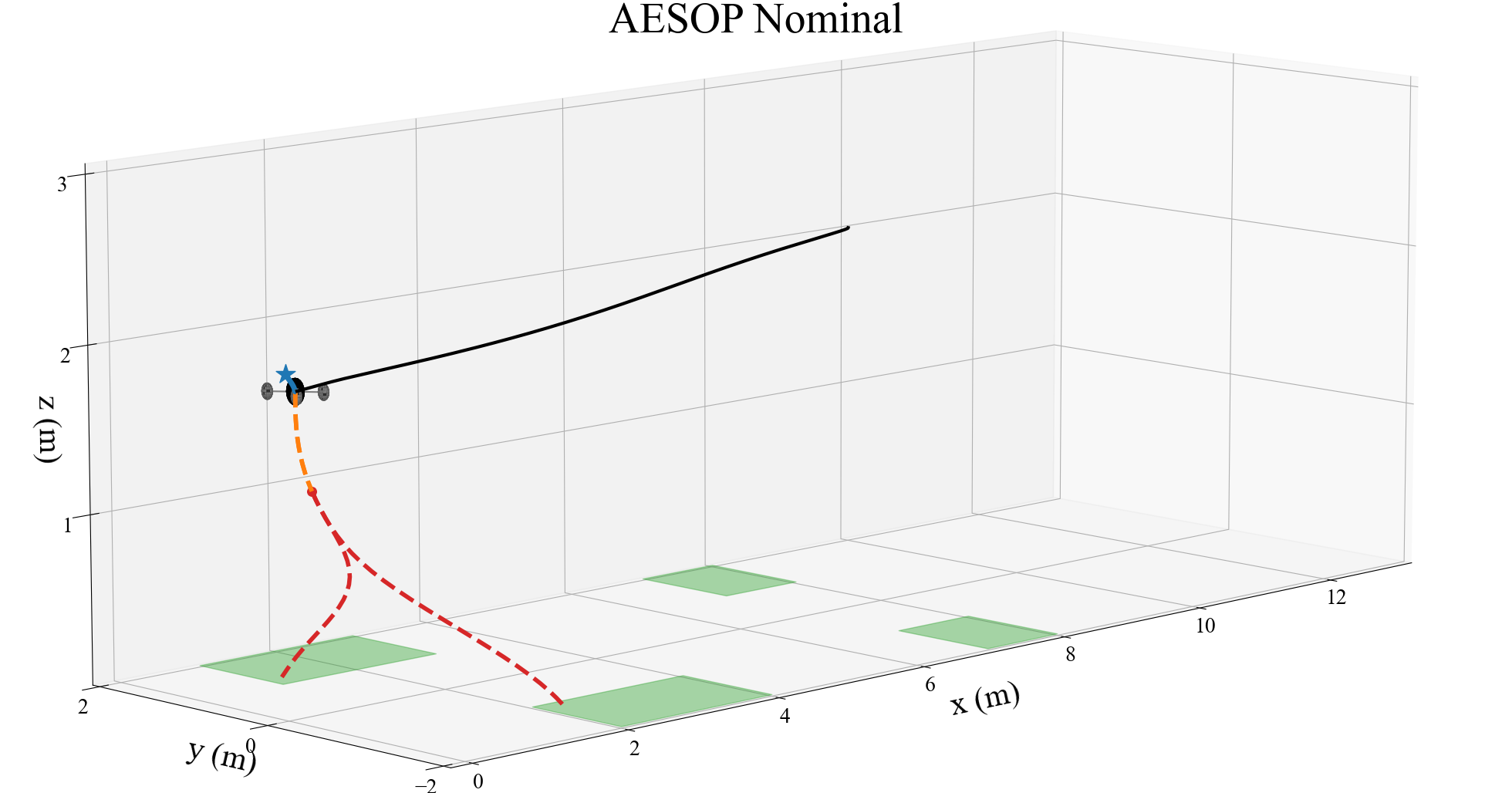}
        \caption{Closed loop trajectory of the AESOP algorithm in a nominal episode (black). That is, an episode in which the fast anomaly detector detects no anomalies. Also shown are the predicted trajectories at the last timestep of the episode.}
        \label{fig:aesop-nom}
    \end{subfigure}
    \hfill
    \begin{subfigure}[t]{0.47\textwidth}
    \includegraphics[width=\textwidth]{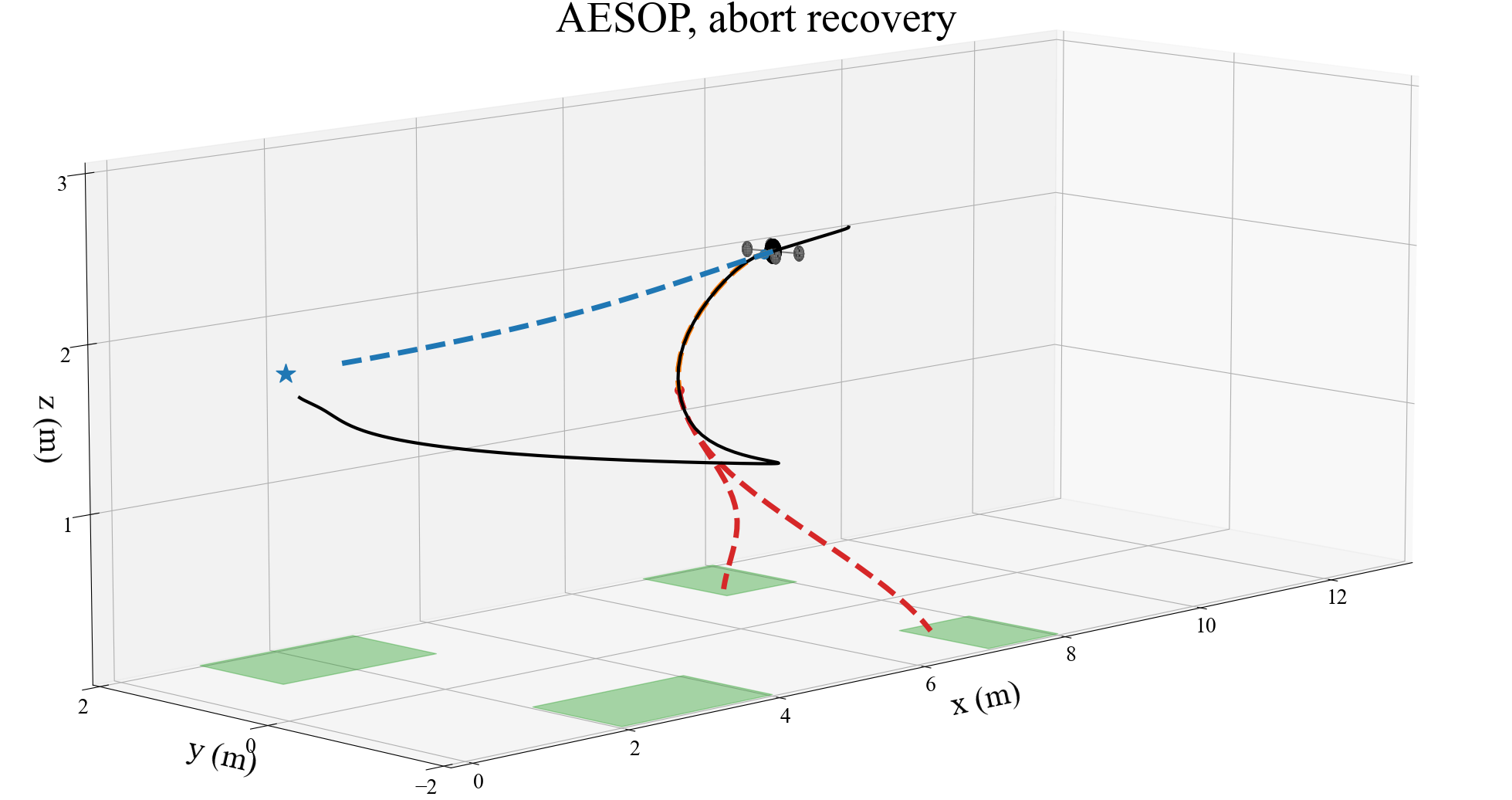}
        \caption{Closed loop trajectory of the AESOP algorithm (black). In this trajectory, the fast anomaly detector signals that an anomaly has been detected. Then, the Slow LLM reasoner outputs that the anomaly is inconsequential to the robot's safety, thereby returning the AESOP algorithm to nominal operation. In blue and red we show the respective nominal and recovery plans computed at the timestep that the fast anomaly detector issues a warning.}
        \label{fig:aesop-abort}
    \end{subfigure}
    \vspace{15pt}
    \\
    \begin{subfigure}[t]{0.47\textwidth}
    \includegraphics[width=\textwidth]{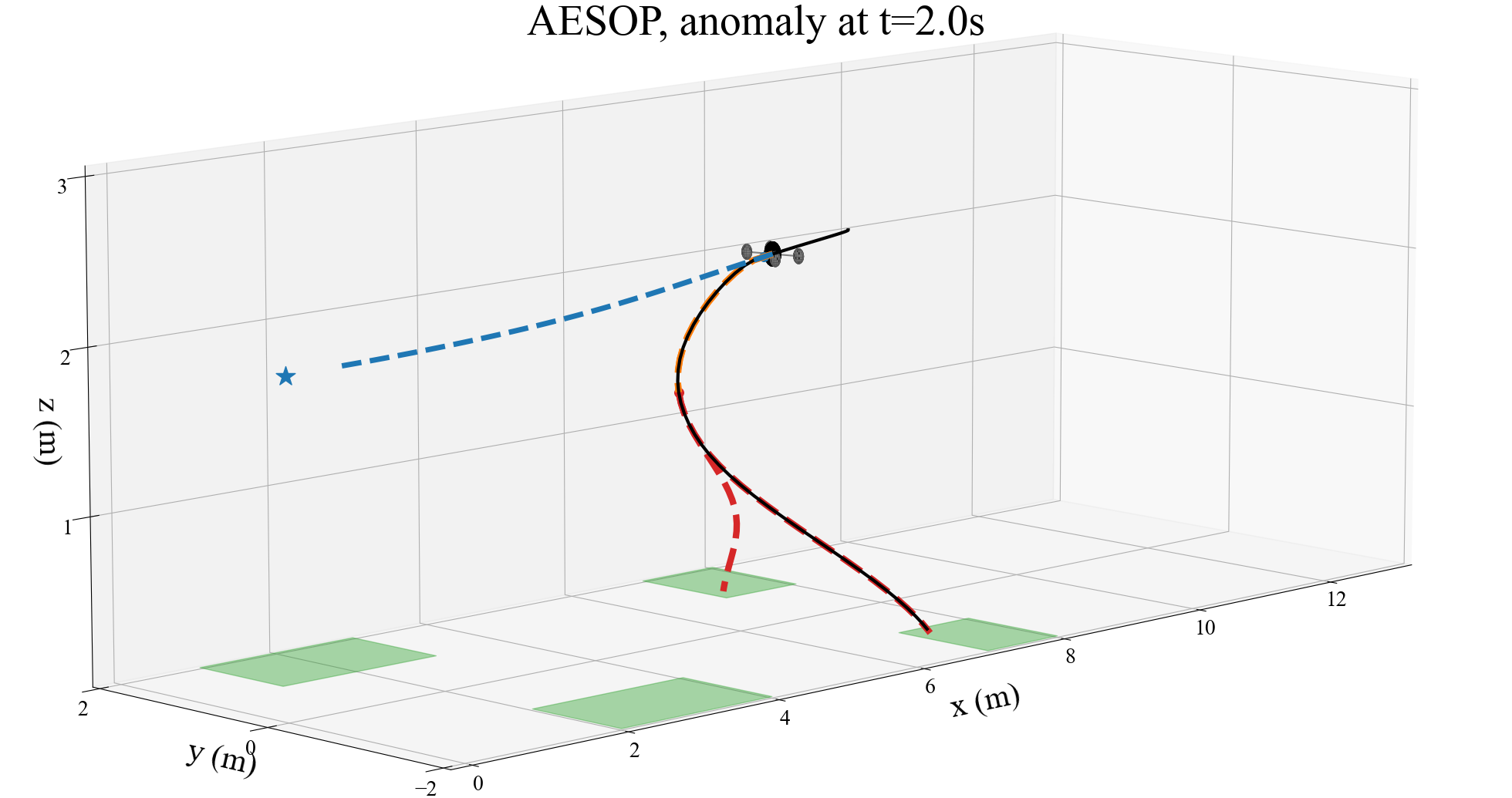}
        \caption{Closed loop trajectory of the AESOP algorithm (black). In this trajectory, the fast anomaly detector signals that an anomaly has been detected at $t=2.0\mathrm{s}$. Then, the Slow LLM reasoner selects the appropriate recovery set from the available options. In blue and red we show the respective nominal and recovery plans computed at the timestep that the fast anomaly detector issues a warning.}
        \label{fig:aesop=early}
    \end{subfigure}
    \hfill
    \begin{subfigure}[t]{0.47\textwidth}
    \includegraphics[width=\textwidth]{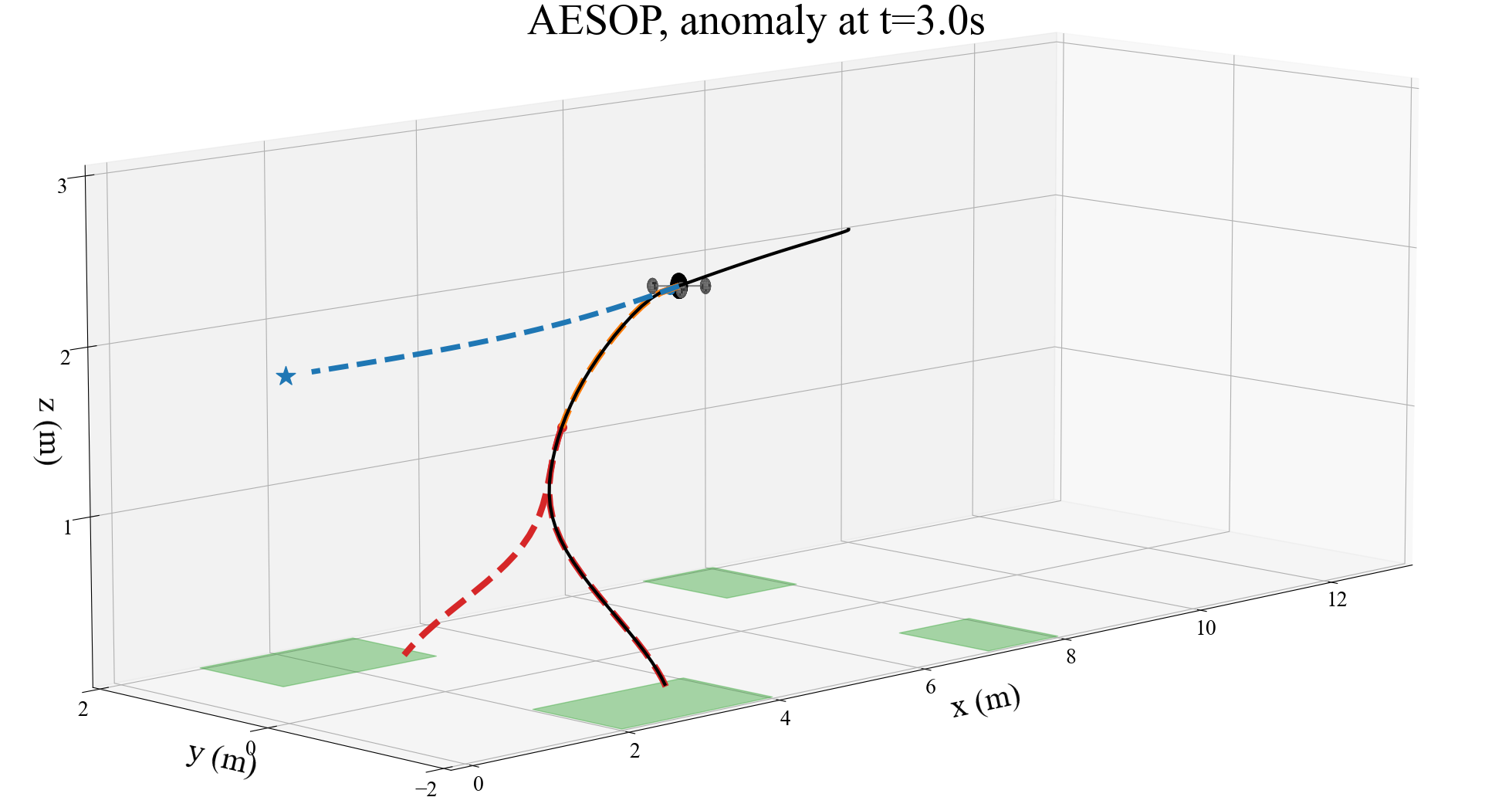}
        \caption{Closed loop trajectory of the AESOP algorithm (black). In this trajectory, the fast anomaly detector signals that an anomaly has been detected at $t=3.0\mathrm{s}$. Then, the Slow LLM reasoner selects the appropriate recovery set from the available options. In blue and red we show the respective nominal and recovery plans computed at the timestep that the fast anomaly detector issues a warning.}
        \label{fig:aesop-late}
    \end{subfigure}
    \vspace{2em}\\
    \caption{Closed loop trajectories of 1) the naive MPC baseline, 2) the Fallback-Safe MPC baseline, 3) the AESOP algorithm in various scenarios.}
    \label{fig:aesop-compare}
\end{figure*}

We simulate the full 12-state dynamics of a quadrotor with an arm length of $.25\mathrm{m}$, a mass of $1\mathrm{kg}$ and an inertia matrix $J = \mathrm{diag}(.45, .45, .7)$. In contrast with the simulation in \cref{sec:expts}, where the velocity of the quadrotor was unconstrained, we constrain the drone to move at a maximum of $1.5 \mathrm{m/s}$ along each of the principle axes' directions. As shown in green in \cref{fig:aesop-compare}, we provide the planner with four recovery regions, encoded as polytopic constraints on states, representing potential landing zones. We label the first two, around $x \approx 8\mathrm{m}$, as grassy fields. We label the third and fourth landing zones around $x \approx 3\mathrm{m}$ as a building rooftop and a parking lot respectively. We use these labels to simulate the EVTOL synthetic in closed loop, abstracting perception into a pure-text observation at each timestep. The quadrotor plans trajectories of length $T=4\mathrm{s}$ using a time discretization of $\mathrm{dt} = .1\mathrm{s}$ and we assume the slow LLM reasoner takes at most $K=1.5\mathrm{s}$ to output a decision. In these simulations, the goal of the quadrotor is to fly to a goal point, denoted by the blue star in \cref{fig:aesop-compare}.

We implement the MPC \cref{eq:modification-mpc} core to the AESOP algorithm in Python using the OSQP \cite{osqp} solver using linearized dynamics and a quadratic objective. Note that AESOP (Algorithm \ref{alg:safe-mpc}) requires a methodology to select a set of recovery regions to constrain the MPC \cref{eq:modification-mpc} at each timestep, and the MPC is guaranteed to be feasible for at least one such choice by \cref{thm:safety}. In principle, one could select the optimal subset of recovery regions using mixed-integer programming. Here, we adopt the simpler approach proposed in \cite{SinhaSchmerlingEtAl2023} where we solve multiple versions of \cref{eq:modification-mpc} at each timestep, each associated with a different combination of at least two recovery sets. Then, we select the trajectory plan associated with the feasible solution to \cref{eq:modification-mpc} with least cost. This allows the quadrotor to dynamically select various safety interventions to maximally make progress towards the goal. 

Overall, this implementation runs at approximately $42 \mathrm{Hz}$ on the Nvidia Jetson Orin AGX's CPU, with some variance induced by re-initialization of solver warm-starts when the recovery sets change. Combined with the inference latency of the fast anomaly detectors in \cref{tab:jetson-timings}, this means that the AESOP framework can comfortably run in real-time.

\subsubsection{Baselines}
We compare AESOP to two baselines that also use the slow LLM-based reasoner to select a recovery set on cue of the fast anomaly detector.

\textbf{Naive MPC:} The first is a naive MPC algorithm that only plans a single nominal trajectory towards the goal during nominal timesteps (i.e., when $\h(\obs_t) = 0$) without maintaining feasible recovery plans. This baseline continues nominal operation until the LLM returns a choice of recovery set, at which point the MPC attempts to plan a recovery trajectory.

\textbf{Fallback-Safe MPC (FS-MPC) \cite{SinhaSchmerlingEtAl2023}:} We base the second baseline on the Fallback-safe MPC proposed in \cite{SinhaSchmerlingEtAl2023}. This algorithm maintains several feasible recovery trajectories at all times in a similar fashion to the AESOP algorithm. However, this algorithm does not account for the latency associated with the slow LLM-based reasoner (i.e., setting $K = 0$ in \cref{eq:modification-mpc}), and only engages a recovery plan once the LLM returns a decision. 

Our implementations of both these algorithms rely on slack variables, so that they return a trajectory plan that minimally violates constraints in case it is dynamically impossible to compute a recovery trajectory that reaches the chosen set. 
\subsubsection{Results} 
\textbf{Qualitative Figures:} In \cref{fig:aesop-compare}, we show the qualitative differences in the behavior of AESOP and the baseline methods. 

Firstly, \cref{fig:aesop-nom} shows the trajectory of the quadrotor using AESOP in an episode where no anomalies occur and therefore, where the fast anomaly detector raises no alarms. As such, \cref{fig:aesop-nom} shows that AESOP does not interfere significantly with the nominal operation of the quadrotor: It still reaches the goal location with little impediment.

Secondly, \cref{fig:naive-mpc} shows the trajectory of the naive MPC baseline in an episode where the fast reasoner detects an anomaly at $t=3.0\mathrm{s}$ and the slow LLM reasoner returns it's output $1.5\mathrm{s}$ later. The naive MPC assumes that all the recovery regions are always reachable from the current state, nor does it account for the latency of the LLMs reasoning. Therefore, once the LLM outputs that the quadrotor should land at a grassy recovery region (around $x=8\mathrm{m}$ in \cref{fig:aesop-nom}), it can no longer plan a dynamically feasible path to the recovery set and crash lands in an unsafe ground region. In contrast \cref{fig:aesop-late} shows that AESOP recovers the robot to the desired safe region. This example showcases that it is necessary to plan multiple trajectories even in nominal scenarios to ensure that the safety-preserving interventions selected by the LLM can be executed safely.

Third, \cref{fig:fsmpc} shows the trajectory of the fallback-safe MPC (FSMPC) algorithm from \cite{SinhaSchmerlingEtAl2023} on the same example as \cref{fig:naive-mpc} and \cref{fig:aesop-late}. While the FSMPC algorithm maintains several feasible recovery plans during nominal operations, \cref{fig:fsmpc} shows that this is not sufficient to ensure safe recovery: The FSMPC algorithm does not account for the time delay of the LLM-based reasoner. Instead it continues it's nominal operations until the LLM returns. As shown in \cref{fig:fsmpc}, the feasible recovery strategies changed in the time interval between the detection of the anomaly detector and the output of the LLM. As a result, the FSMPC algorithm tried to find a dynamically feasible recovery trajectory to the grassy areas as instructed by the LLM, but was instead forced to crash land in an unsafe region. In contrast, by accounting for the latency of the LLM, AESOP was able to safely land (i.e., \cref{fig:aesop-late}), thus showing the necessity of accounting for LLM inference latency in control design for dynamic robotic systems.

Finally, \cref{fig:aesop-abort} shows the behavior of the AESOP algorithm when the fast anomaly detector detects an anomalous scenario, but the LLM determines the anomaly is inconsequential to the robot's safety and returns the system to nominal operations. The quadrotor descends and slows down during the inference time of the LLM, and continues its flight towards the goal thereafter. As such, \cref{fig:aesop-abort} shows that by leveraging the LLM, AESOP minimally impedes goal completion when anomalies are not immediate safety risks. Moreover, \cref{fig:aesop-late} and \cref{fig:aesop=early} show that AESOP safely recovers the system when the anomaly is consequential to safety.

\textbf{Quantitative Evaluation:}
We quantitatively ablate the improvement of our proposed approach in comparison with the naive MPC and FSMPC by simulating $500$ trajectories with a consequential anomaly appearing at a random timestep. To do so, we uniformly sample an initial condition with zero velocity and rotation in a box with width $2\mathrm{m}$ around $(x,y,z) = (10,2,2)$ and fix the goal state as in \cref{fig:aesop-compare}. We then uniformly select a timestep at which the anomaly appears in the interval $t=[1\mathrm{s}, 4\mathrm{s}]$. \cref{tab:drone-ablate} shows the fraction of scenarios in which AESOP and the baselines safely reach the recovery set chosen by the LLM. By design, AESOP successfully lands the quadrotor in the chosen recovery region each time. While the FSMPC algorithm improves over the naive baseline, \cref{tab:drone-ablate-app} shows that it is essential to account for the LLM's latency to achieve reliability.

\begin{table}[]
\centering
\caption{Percentage of trajectories where the quadrotor successfully recovered to the LLM's choice of recovery region.}
\label{tab:drone-ablate-app}
\begin{tabular}{p{2cm}ccc}
\toprule
 & Naive MPC & FS-MPC \cite{SinhaSchmerlingEtAl2023} & AESOP \\ 
\midrule
Successful Recovery Rate &  15\% &  23\% & \textbf{100\%} \\
\bottomrule
\end{tabular}
\end{table}

\subsection{Additional Results: Autonomous Vehicle Simulation}\label{sec:ablation}

\begin{figure}[b]
    \centering
    \begin{subfigure}{0.9\textwidth}
        \includegraphics[width=\linewidth]{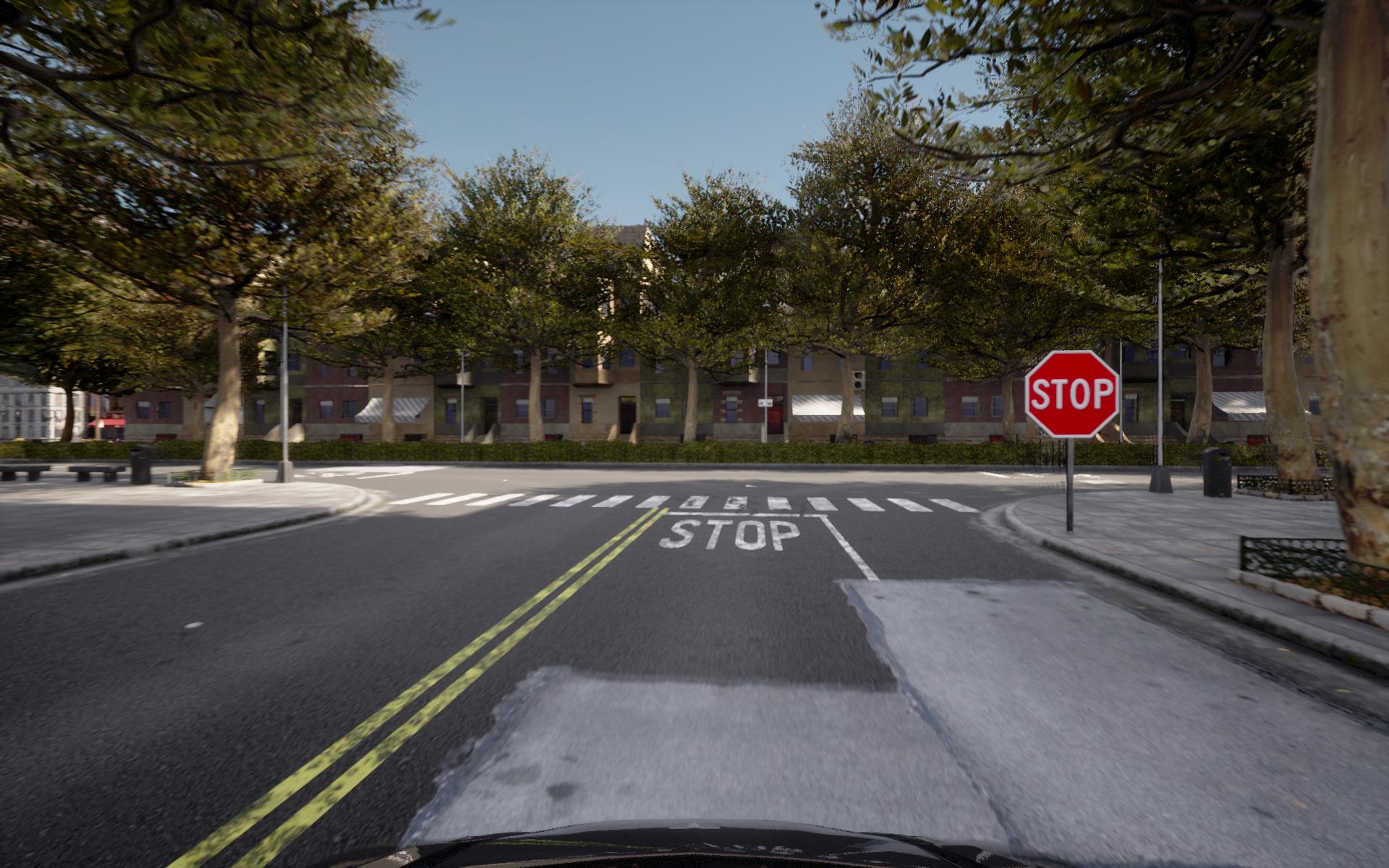}
        \caption{Vehicle approaches a nominal stop sign.}
        \label{fig:nom-ss}
    \end{subfigure}
    
    \bigskip
    \smallskip
    
    \begin{subfigure}{0.9\textwidth}
        \includegraphics[width=\linewidth]{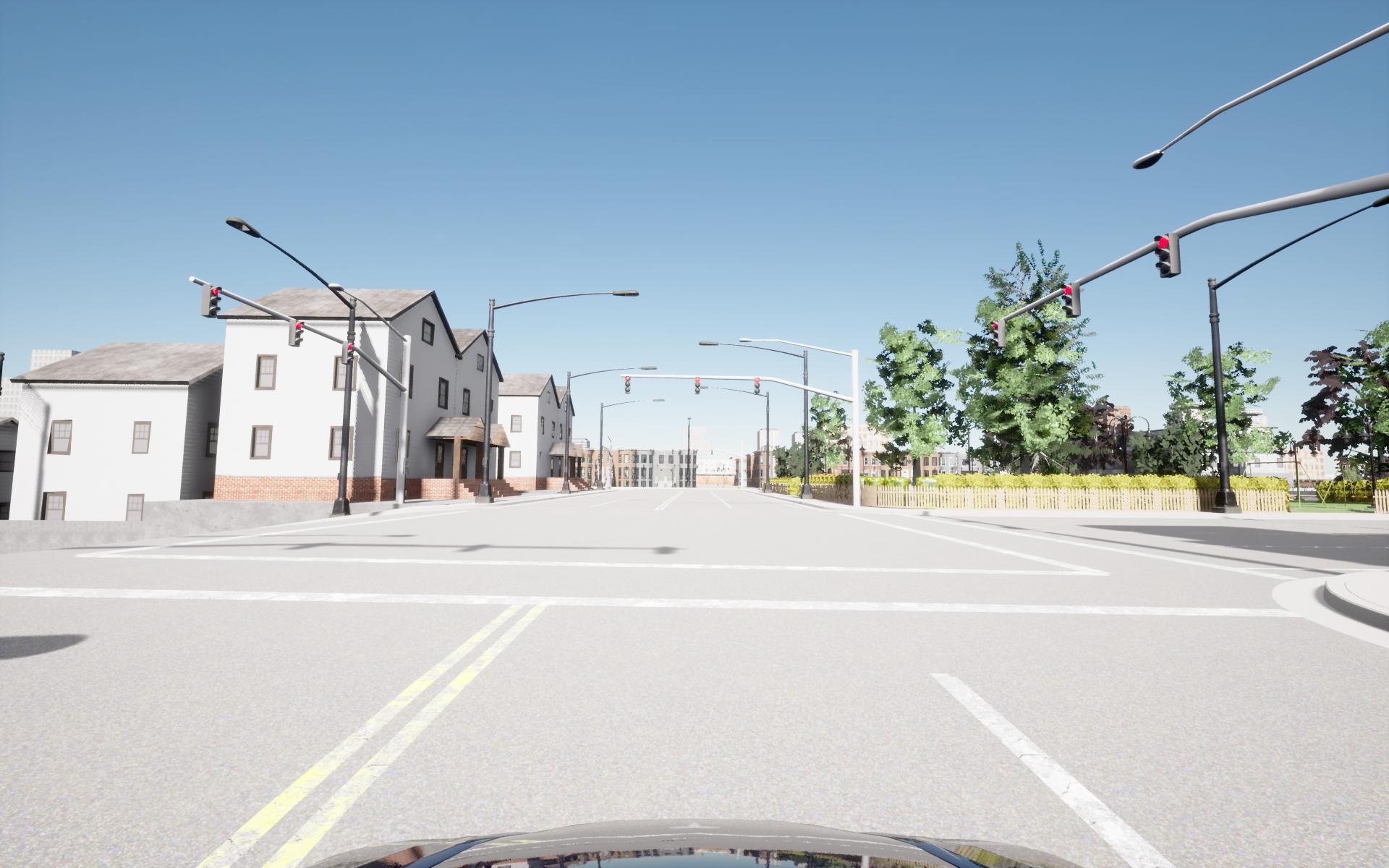}
        \caption{Vehicle approaches a nominal traffic light.}
        \label{fig:nom-tl}
    \end{subfigure}
    
    \smallskip
    
    \caption{Nominal observations for each object class in \cite{elhafsi2023}.}
\end{figure}

\begin{figure}
    \centering
    \begin{subfigure}{0.9\textwidth}
        \includegraphics[width=\linewidth]{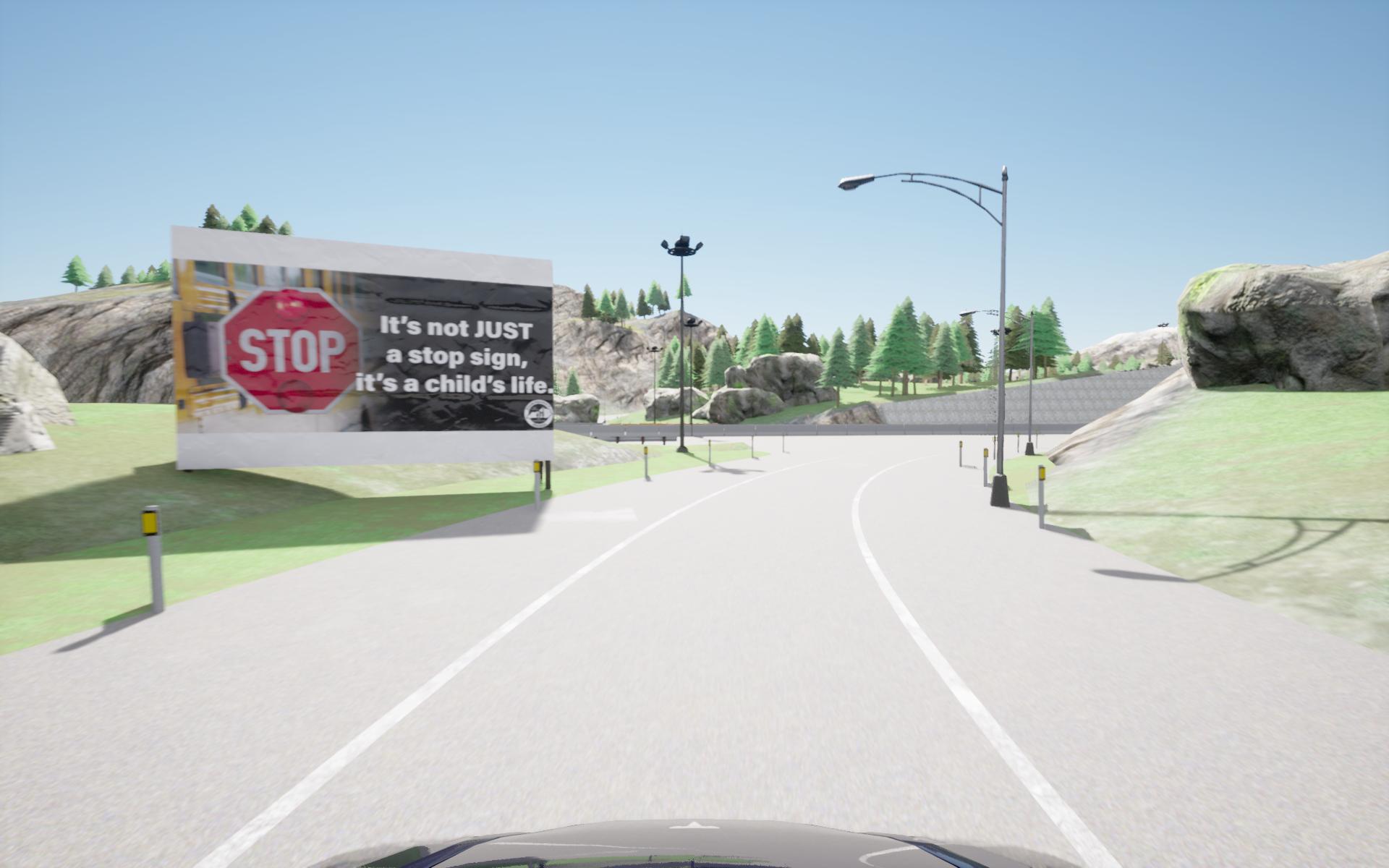}
        \caption{Vehicle approaches a stop sign anomaly.}
        \label{fig:anom-ss}
    \end{subfigure}
    
    \bigskip
    \smallskip
    
    \begin{subfigure}{0.9\textwidth}
        \includegraphics[width=\linewidth]{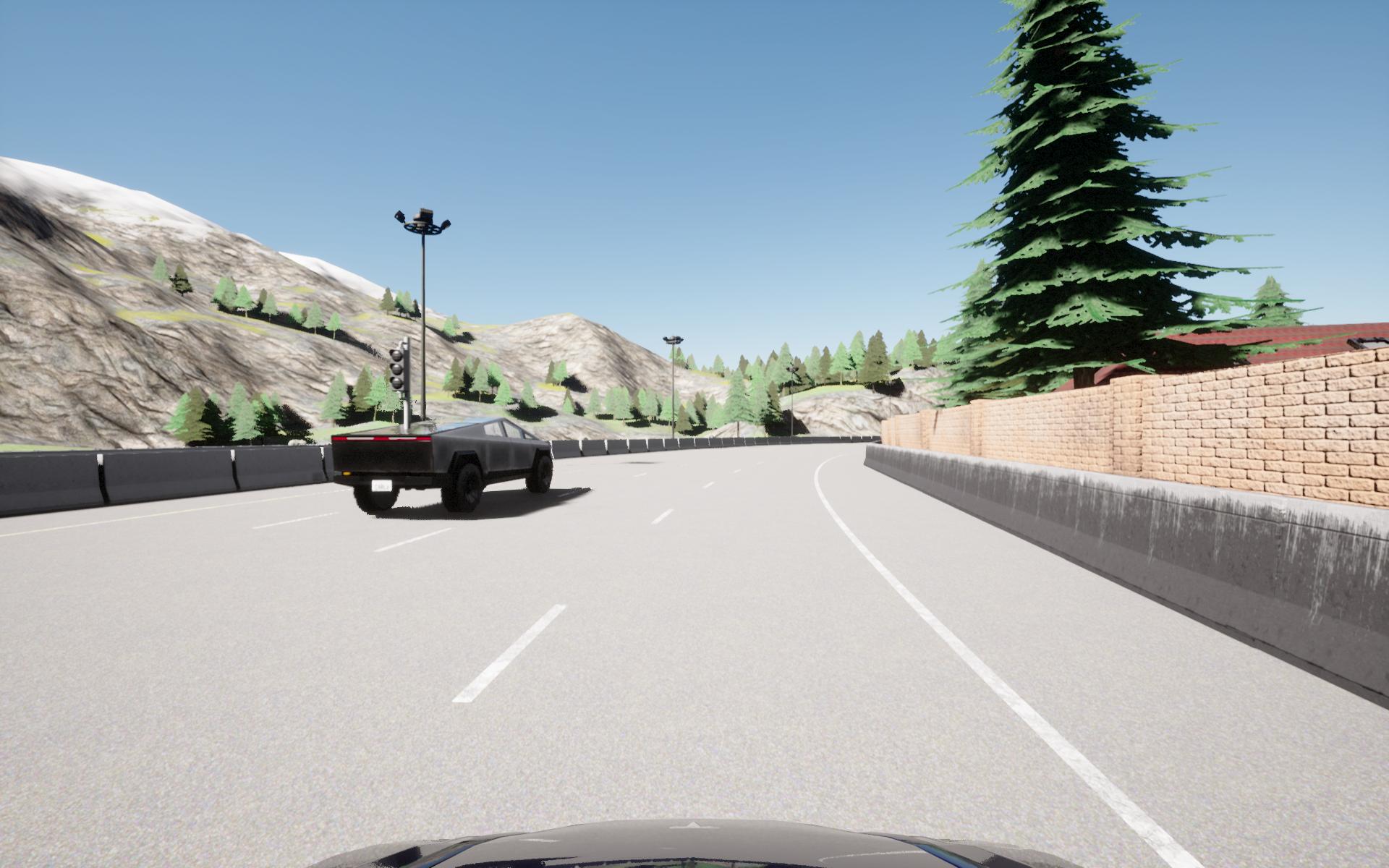}
        \caption{Vehicle approaches a traffic light anomaly.}
        \label{fig:anom-tl}
    \end{subfigure}
    
    \smallskip
    
    \caption{Anomalous observations for each object class in \cite{elhafsi2023}.}
\end{figure}

In our ablations on self-driving scenarios, we adopt the semantic anomaly dataset presented in \cite{elhafsi2023}. This dataset includes two classes of semantic anomalies with multiple instantiations in a set of independent experiments. In nominal experiments, the vehicle approaches a stop sign, as in Fig. \ref{fig:nom-ss}, or a traffic light, as in Fig. \ref{fig:nom-tl}, in an environment with common-place observations. In anomalous experiments, the vehicle approaches an image of a stop sign on a billboard, as in Fig. \ref{fig:anom-ss}, or a truck transporting an inactive traffic light, as in Fig. \ref{fig:anom-tl}. In this setting, we aim to use observations gathered from trajectories including nominal stop signs and traffic lights to identify when either anomaly is present in a novel observation. We adapt our anomaly detection pipeline using a two-step approach with language embeddings and a direct end-to-end method with multi-modal CLIP embeddings. 

Supplemental to our discussion in the paper's main body, our ablations provide four interesting observations: 1) ground truth scene descriptions are necessary to overcome misclassifications by the object detector, 2) a large model, on the scale of Mistral (7B), is necessary to capture the presence of an anomaly in a semantically rich environment such as self-driving, 3) embeddings from the two-stage pipeline using LM embeddings only capture semantics and do not pick up on visual novelty, whereas multi-modal CLIP embeddings incur false positives on visually novel but semantically nominal observations, and 4) that a smaller model, like MPNet, struggles to correctly detect anomalies as the number of objects in an observation increases.

\subsubsection{Object Detection vs. Ground-truth Detections}
\begin{figure}
    \centering
    \includegraphics[width = \textwidth]{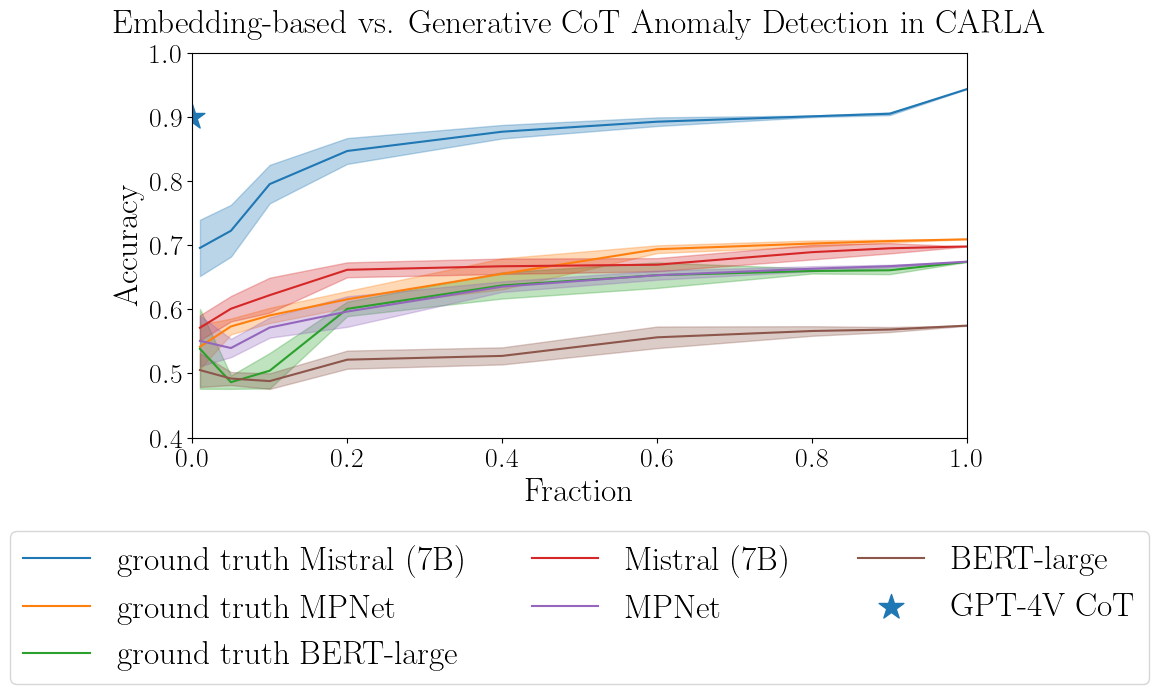}
    \caption{Anomaly detection with MPNet, BERT-large and Mistral (7B) w/ and w/out ground truth scene descriptions against generative CoT reasoning with GPT-4V.}
    \label{fig:carla-language-embeddings}
\end{figure}

Recall, the two-step pipeline constructs a textual prompt using scene descriptions output by OWL-ViT on each CARLA observation. Our initial experimentation found processing language embeddings from the raw scene descriptions returned by OWL-ViT was insufficient to surpass GPT-4V's CoT accuracy which is demonstrated in Fig. \ref{fig:carla-language-embeddings}. As noted in \cite{elhafsi2023}, because CARLA's synthetic visual features represent a distribution shift from the realistic images on which OWL-ViT was trained, OWL-ViT periodically hallucinates object detections, such as the presence of an anomaly when none is present, or misses an anomaly detection entirely. Most commonly, we noticed the OWL-ViT characterized the rendered images as e.g., ``an image of a stop sign'' or ``a picture of a stop sign.'' Therefore, independent of the model's size, the language embeddings on anomalous scene descriptions are not significantly different from that of nominal observations: at best, we achieve 0.70 mean accuracy with Mistral (7B) which is inferior to GPT-4V CoT. Instead, if we post-process the raw scene descriptions to contain ground-truth detections, then the resultant language embeddings are semantically different between the nominal and anomalous observations. We do this by removing false positive detections and introducing a true positive detection if missed by the detector. We demonstrate this finding in Fig. \ref{fig:carla-language-embeddings}, which shows Mistral (7B) achieves a mean accuracy of 0.94 surpassing GPT-4V CoT by 4\% absolute. While post-processing an object detector's output is not feasible at deployment, this ablation serves as a proof of concept for our algorithm in a real-world environment where object detection is presumably reliable. 

Also, we notice that when using ground truth scene descriptions, the model's size creates a spread in performance in Fig. \ref{fig:carla-language-embeddings}. Interestingly, with the introduction of ground truth detections, MPNet  and BERT-large achieve similar accuracy to Mistral (7B) processing raw scene descriptions. With Mistral (7B), which is 64x and 19x larger than MPNet and BERT-large, respectively, access to ground truth detections completely unblocks anomaly detection. These results suggest that MPNet and BERT-large, with limited expression due to model size, produce language embeddings that are not semantically rich enough to capture the presence of an anomaly among numerous nominal objects. Therefore, our two-step pipeline requires high-fidelity scene descriptions and a sufficiently large language model, on the scale of Mistral (7B), for anomaly detection with high-dimensional observations characteristic of the real world. 

\subsubsection{Calibration Ablation on Withheld Trajectories}
\begin{table}[b]
    \centering
    \caption{Accuracy of embedding detectors when withholding nominal data from CARLA routes with anomalies. All methods are our own.}
    \label{tab:carla-withheld-ablation}
    \begin{tabular}{clccc}
        \toprule
        \textbf{} & Method & TPR & FPR & Bal. Accuracy \\
        \midrule
            & (Lang.) MPNet Abl. & 0.55 & 0.11 & 0.72 \\
            & (Lang.) Mistral Abl. & 0.96 & 0.19 & 0.89\\
        \midrule
            & (Vision) CLIP Abl. & 0.99 & 0.57 & 0.71\\
        \bottomrule
    \end{tabular}
\end{table}
The image observations in the CARLA dataset from \cite{elhafsi2023} were constructed by driving a car along simulated routes in several different maps. Some of these routes pass by anomalous objects, e.g., a stop sign on a billboard, which trick the vehicle into making unsafe decision. This means that episodes wherein the vehicle takes unsafe actions include both nominal and anomalous observations, and that the routes appear both with and without anomalous objects. In the main evaluations in \cref{tab:carla-anomaly-detection} and \cref{sec:ablation} the embedding caches therefore contain nominal embeddings associated with all the routes, i.e., capturing a setting wherein anomalies suddenly appear on roads the AV often drives. As shown in \cref{tab:carla-anomaly-detection}, the two-stage detectors (using MPNet and Mistral) and the single stage multi-modal detector (using CLIP) perform well at detecting anomalies in this scenario. 

Therefore, we also run an ablation wherein we withhold all nominal data from routes wherein anomalies occur when constructing the embedding cache, leaving only routes in which anomalies never occur. This resembles a setting where the vehicle drives novel routes that contain sporadic anomalies. As shown in \cref{tab:carla-anomaly-detection}, the false positive rate of the CLIP embedding-based approach significantly increases in this scenario. This is most likely because the CLIP embeddings contain both visual features (e.g., those suitable for object detection as used in \cite{sun2023simple}) and semantic features. Therefore, the visual novelty in the previously unseen trajectories causes false positives. In contrast, the two-stage approaches, which first describe the scene with an object detector before prompting an LM embedding model, do not attend to visual features and therefore do not suffer such performance drops. In \cite{KaramchetiEtAl2023}, the authors argue that multimodal models for robotics should explicitly balance visual and semantic features, and in line with those insights, we argue future work investigating how to differentiate visual and semantic features can make multi-modal anomaly detectors more robust.

\subsubsection{Accuracy vs. Complexity of Observations}
\begin{figure}[t]
    \includegraphics[width=\textwidth]{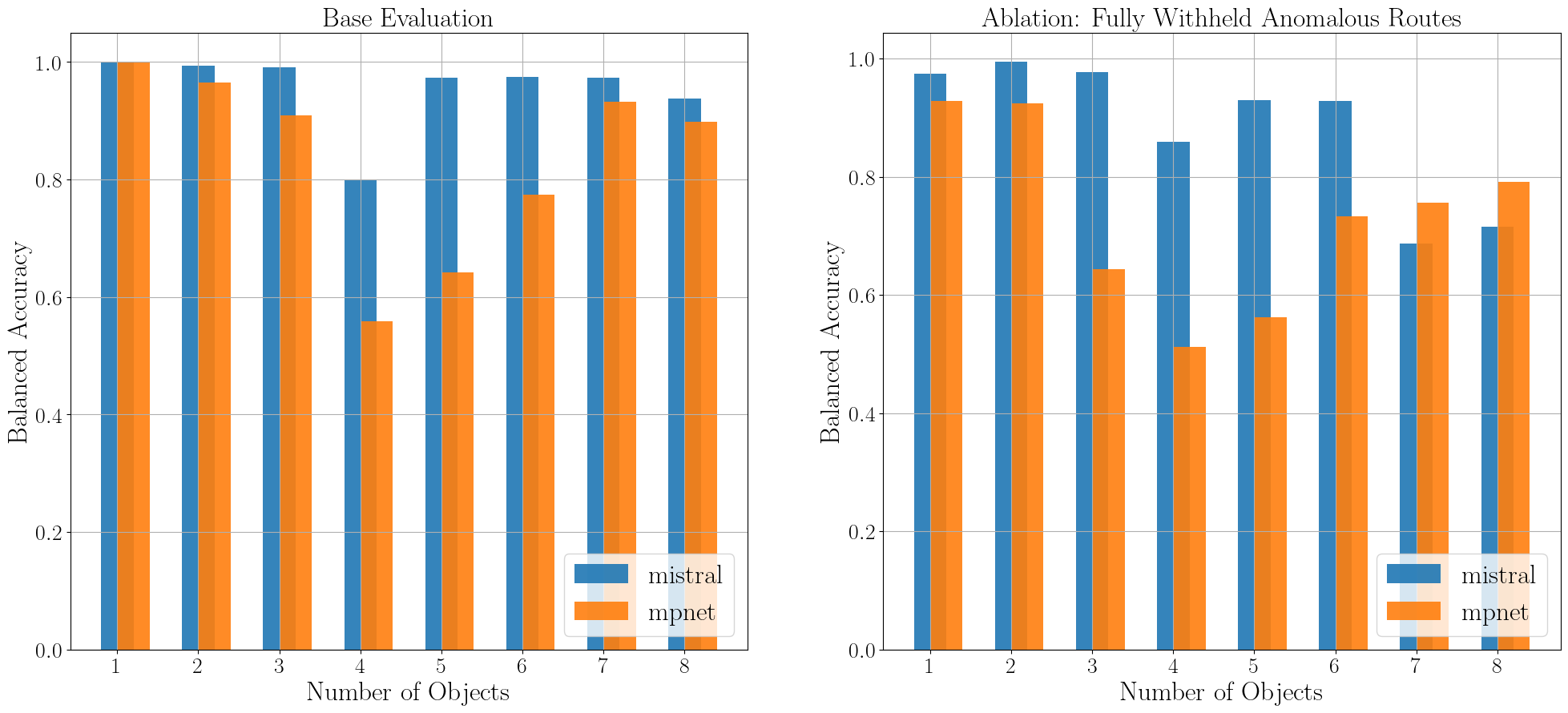}
    \caption{Accuracy of the text-based embedding detectors as a function of the number of objects within each observation for the CARLA dataset.}
    \label{fig:acc_with_detections}
\end{figure}

\begin{figure}[b]
    \includegraphics[width=.8\textwidth]{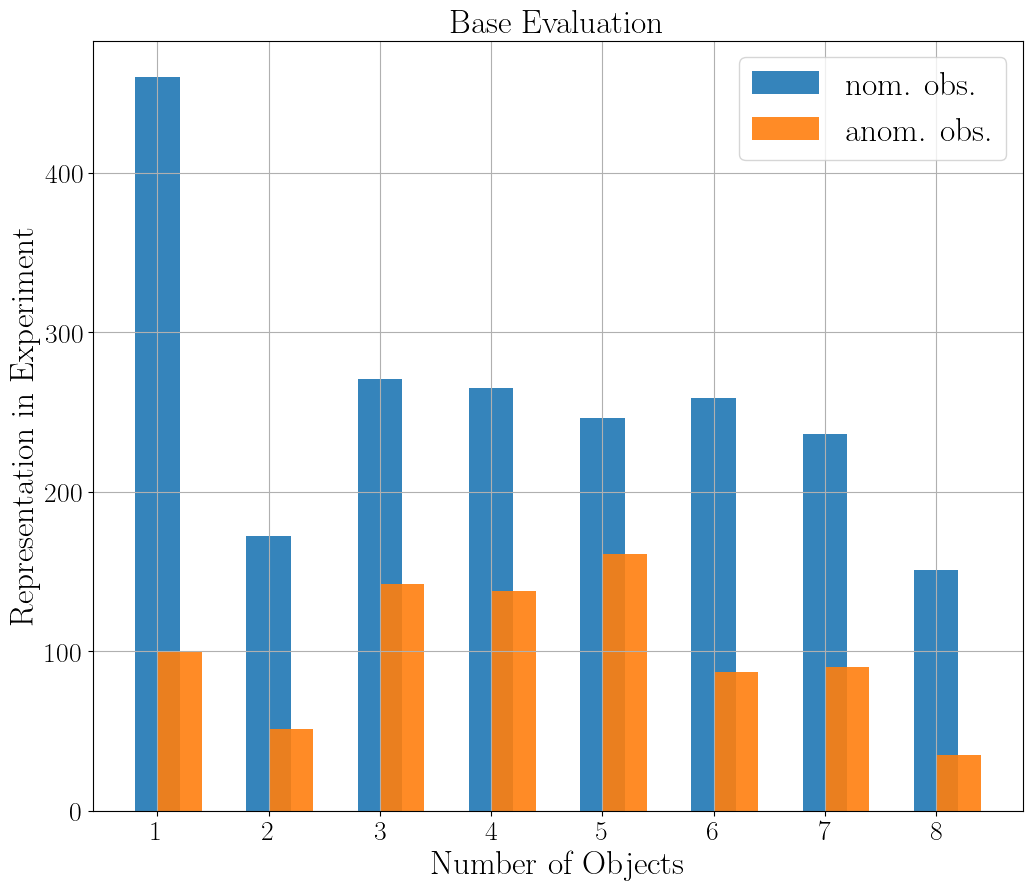}
    \caption{Representation of nominal (nom.) and anomalous (anom.) observations (obs.) in the our CARLA evaluation dataset.}
    \label{fig:obs_representation}
\end{figure}

Here, we further investigate the discrepancy in accuracy between the smaller MPNet embedding model (110M) and the larger Mistral embedding model (7B) in \cref{tab:carla-anomaly-detection}. We do so because on the synthetic tasks in \cref{sec:exps-synthetics}, where the observations contain descriptions of at most three objects, both models performed more comparably. As shown in \cref{fig:acc_with_detections}, we see that the accuracy of MPNet drops off when the total number of objects within each image observation increases, which largely explains their difference in performance. However, an interesting nuance is that MPNet's accuracy is again comparable to Mistral when there are 7-8 objects in the observation. \cref{fig:obs_representation} supports the hypothesis that this may be because the imbalance between nominal and anomalous images is larger for observations with many objects, so that large numbers of observations may correlate with nominal conditions. Moreover, there are significantly fewer observations with many obstacles. Overall, these results suggest that larger models are needed to reason about the anomalousness of more complex scenes with many objects.

\subsection{Prompting Details}\label{ap:prompting}
We emphasize that all the prompts used in our experiments can be found in the repositories listed on our project webpage, \texttt{\url{https://sites.google.com/view/aesop-llm}}. However, for completeness, we briefly describe our prompting strategy here. For all language-based tasks (i.e., the synthetics in \cref{sec:exps-synthetics}, the hardware in \cref{sec:exps-hardware}, and the text-based evaluations in \cref{sec:exps-ablation}), we parse the current task description of the robot and a list of all observed objects into a prompt template. The template first provides a brief description of the robot that is being monitored and defines the monitoring task, followed by the current task and observation. The prompt finally concludes with an instruction on the monitor's output (e.g., a single token nominal/anomaly classification or a chain-of-thought safety assessment). For multi-modal models, we only provide the robot's task. In early experiments, we found that it was necessary to provide the descriptions of the robot and the monitoring task so that the model interprets the observations from the context of the robot's task, whereas e.g., simply embedding an object label or asking an LLM whether an object is anomalous fails to capture its relation to the robot and other objects in the environment. As a concrete example, we use the following template for the synthetic VTOL task in \cref{sec:exps-synthetics}:
\begin{mdframed}[backgroundcolor=light-gray, roundcorner=10pt,leftmargin=1, rightmargin=1, innerleftmargin=15, innertopmargin=15,innerbottommargin=15, outerlinewidth=1, linecolor=light-gray]
\begin{lstlisting}[linewidth=\columnwidth,breaklines=true]
"I am the runtime monitor for a vision-based autonomous vertical takeoff and landing (VTOL) aircraft operating as an urban air taxi. My job is to analyze the observations of the VTOL and identify potential anomalous objects that the VTOL may be unable to handle as part of its nominal operations. The VTOL's observations take the form of object detections generated by an onboard perception pipeline.\n

Scene Description:\n
The VTOL's current task is to {TASK}.\n
The VTOL's current observation is:\n
{OBSERVATION}\n
I will reason about whether the VTOL should be able to handle the observation as part of its nominal operations. Do any of the detected objects qualify as anomalies beyond the standard operating conditions and capabilities of the VTOL?"
\end{lstlisting}
\end{mdframed}

\subsection{Proof of \cref{thm:safety}}\label{supp:theorem-proof}
Finally, we prove prove \cref{thm:safety}, which establishes the properties of the closed loop system formed by AESOP (Algorithm \ref{alg:safe-mpc}) and \cref{eq:dyn}. To do so, the assumption that $\calX_R^1, \dots, \calX_R^d$ are control invariant sets, which we made in the problem formulation (see \cref{sec:problem-formulation}), is critical. Therefore, to make this paper as self-contained as possible, we first reiterate the standard definition of a control invariant set. 
\begin{definition}[Control Invariant Set \cite{BorrelliBemporadEtAl2017}]
    A set $\calX_R \subseteq \calX$ for the dynamical system \cref{eq:dyn} subject to state and input constraints $\calX \subseteq \R^n$, $\calU \subseteq \R^m$ is a \emph{control invariant set} if for every $\x \in \calX_R$, there exists a $\fu \in \calU$ such that $\f(\x, \fu) \in \calX_R$. 
\end{definition}

As shown in our experiments in \cref{sec:exps-drone}, \cref{sec:exps-drone}, it is often straightforward to identify recovery regions that are control invariant: For example, states in which the quadrotor has landed are control invariant. We now restate \cref{thm:safety} and provide its proof, which relies on a recursive feasibility argument \cite{BorrelliBemporadEtAl2017}.

\safety*
\begin{proof}
    Suppose that the MPC in \cref{eq:modification-mpc} is feasible for some set of recovery strategies $\calY \subseteq \{1, \dots, d\}$ at some time step $t < t_{\mathrm{anom}}$. Let $\x_{t:t+T+1|t}^{i, \star }$ denote optimal predicted trajectories and let $\fu_{t:t+T|t}^{i, \star}$ be the optimal predicted input sequences associated with a) the nominal trajectory, $i=0$, and b) recovery strategies, $i\in \calY$, that minimize \cref{eq:modification-mpc} at time $t$. Then, it holds that $\x_t \in \calX$ and $\fu_t = \fu_{t|t}^{0,\star} \in \calU$ by construction of \cref{eq:modification-mpc}. 

    Furthermore, since we assume that each recovery set $\calX_R^i$ is a control invariant set, there exists an input $\fu_{t+T+1|t+1}^i \in \calU$ such that the input sequence $\fu_{t+1:t+T+1|t+1}^i := [\fu_{t+1:t+T|t}^{i, \star}; \fu_{t+T+1|t+1}^i]$ and its associated state sequence $\x_{t+1:t+T+2|t+2}^i := [\x_{t+1}; \x_{t+2:t+T+1|t}^{i,\star}; \f(\x_{t+T+1|t}^{i,\star}, \fu_{t+T+1|t+1}^i)]$ satisfy state and input constraints with $\x_{t+T+2|t+1}^i \in \calX_R^i$ for each $i \in \calY$. This implies that 1) there exists a set $\calY' \subseteq \{1, \dots, d\}$ with $|\calY'| \geq 1$ for which the MPC \cref{eq:modification-mpc} is feasible at time $t+1$ and 2) we therefore satisfy $\x_{t+1} \in \calX$ and $\fu_{t+1} \in \calU$. Therefore, we have 1) that $\x_t \in \calX$ and $\fu_t \in \calU$ and 2) there exists at least one safety intervention for which the MPC \cref{eq:modification-mpc} is feasible for all $t \leq t_{\mathrm{anom}}$.

    Next, suppose that the slow reasoner $\w$, queried at $t_{\mathrm{anom}}$, returns an output after exactly $K' \leq K$ timesteps. In addition, suppose that $t_{\mathrm{anom}} \leq t \leq t_{\mathrm{anom}} + K'$, and that the MPC \cref{eq:modification-mpc} is feasible at time $t$ for some set of interventions $\calY \subseteq \{1, \dots, d\}$. 
    Let $k = t - t_{\mathrm{anom}}$. We therefore have that the control and state sequences $\fu_{t+1:t+T - k|t}^{i, \star}$ and $\x_{t+1:t+T+1 - k|t}^{i, \star}$  for $i \in \calY$ are feasible for the MPC \cref{eq:modification-mpc} at $t+1$ using the same set of interventions $\calY$, since Algorithm \ref{alg:safe-mpc} ensures we solve \cref{eq:modification-mpc} with horizon $T - k - 1$ and consensus horizon $K - k - 1$. Since we already proved that \cref{eq:modification-mpc} is feasible at $t_{\mathrm{anom}}$ in the preceding paragraph, it therefore holds by induction that 1) $\x_t \in \calX$ and $\fu_t \in \calU$, 2) that the MPC \cref{eq:modification-mpc} is feasible with respect to the set of feasible safety interventions at time $t_{\mathrm{anom}}$, $\calY_{t_{\mathrm{anom}}}$, for all $ t_{\mathrm{anom}} \leq t \leq t_{\mathrm{anom}} + K'$.

    Now, we consider the case where the slow reasoner outputs $y \in \calY_{t_\mathrm{anom}}$. The preceding step of the proof then shows that there exists a feasible trajectory for the MPC \cref{eq:modification-mpc} that reaches the recovery set output by the LLM, $\calX_R^y$ where $y = \w(\obs_{t_{\mathrm{anom}}})$, within $T + 1 - K$ timesteps. Therefore, by noting that Algorithm \ref{alg:safe-mpc} continues to shrink the prediction horizon, we have that 1) $\x_t \in \calX$ and $\fu_t \in \calU$, 2) that the MPC \cref{eq:modification-mpc} is feasible with respect to the set of recoveries $\{y\}$ for all $t_{\mathrm{anom}} + K' \leq t \leq t_{\mathrm{anom}} + T + 1$. Furthermore, we then also have that $\x_{t_{\mathrm{anom}} + T + 1} \in \calX_R^y$. Because we assume $\calX_R^1, \dots, \calX_R^d$ are control invariant sets, we then further have that the closed loop system formed by \cref{eq:dyn} and Algorithm \ref{alg:safe-mpc} satisfies 1) state and input constraints for all time, 2) that $\x_t \in \calX_R^y$ for all $t \geq t_{\mathrm{anom}} + T + 1$. 

    Finally, we consider the case that the slow reasoner decides to return to nominal operaion, i.e., that $\w(\obs_{t_{\mathrm{anom}}}, \calY_{t_{\mathrm{anom}}}) = 0$. Since each $\calX_R^i$ is a control invariant set, choosing $\calY_{t_{\mathrm{anom}} + K'} = \calY_{t_{\mathrm{anom}}} $ ensures that the MPC \cref{eq:modification-mpc} is feasible at time $t_{\mathrm{anom}} + K'$. The theorem then recursively follows by applying all the preceding steps of the proof.  
\end{proof}

\inlineedit{\subsection{Hardware Experiment -- Additional Details and Results}\label{app:hardware}

\inlineedit{In this section we detail further experimental results that 1) detail our data collection and monitor calibration process, 2) analysis of query latencies to choose the parameter $K$ in \cref{eq:modification-mpc}, 3) quantify the performance of the fast anomaly detector, 4) describe the behavior of the closed-loop system on our test scenarios. Finally, we discuss the implementation details of the MPC controller. Most importantly, we emphasize to the reader that videos of our experiments are included in the supplementary materials, as well as on the \textbf{project web page: \url{https://sites.google.com/view/aesop-llm}}.  

\subsubsection{Data Collection}

As described in \cref{sec:exps-hardware}, we collect data by flying the drone in a circular pattern above the operational area with a nominal clutter of objects on the ground and recording object detections from the observations. We construct a prompt from each possible combination of up to 4 detections from the set of unique detections observed and embed these to form the vector cache. We perform this data collection process twice. First, we collect the embeddings representing the quadrotor's prior experience, $\calD_e$, whereby there are no anomalous elements placed in the scene. We use this dataset to construct the anomaly detector we use in the experiments. In addition, to evaluate the performance of the anomaly detector, we collect a calibration dataset, $\calD_c$, where a limited number of unseen objects are introduced and object placements are varied (e.g., placing objects on top of the box to simulate obstructions). We use $\calD_c$ to evaluate the anomaly detector in the next subsection. We show an example of a nominal and an anomalous observation in \cref{fig:quadrotor-nominal-example} and \cref{fig:quadrotor-anomaly-example}, respectively.

\begin{figure}
    \centering
    \includegraphics[width = \textwidth]{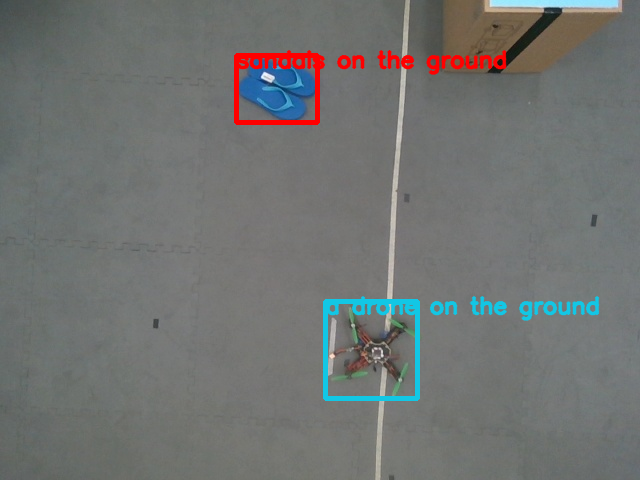}
    \caption{Example of a nominal observation in the quadrotor experiment.}
    \label{fig:quadrotor-nominal-example}
\end{figure}

\begin{figure}
    \centering
    \includegraphics[width = \textwidth]{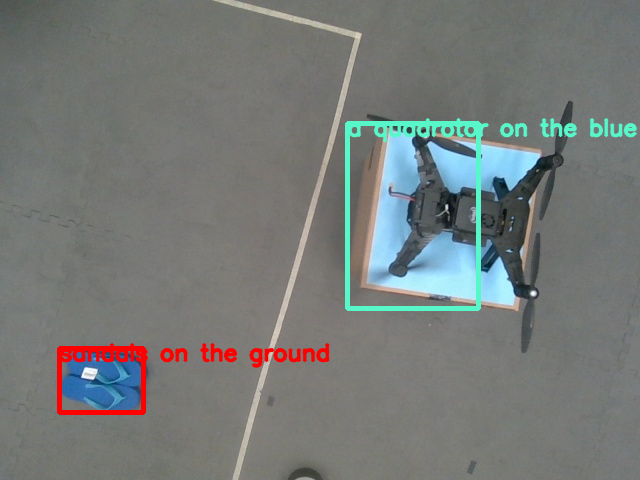}
    \caption{Example of an anomalous observation in the quadrotor experiment due to the obstruction of the landing zone.}
    \label{fig:quadrotor-anomaly-example}
\end{figure}

\subsubsection{Fast Anomaly Detector Calibration}

To calibrate the anomaly detection threshold we compute the top-$k$ score for each embedding in $\calD_c$ against the embedding cache $\calD_p$ and identify the lowest score threshold such that we achieve a TPR of at least 0.9 on the calibration set. We compare the receiver operator characteristic (ROC) curves for top-1 and top-3 scoring in \cref{fig:quadrotor-hardware-calibration}. Interestingly, we find that top-$k$ scoring performed best when $k = 1$, which we attribute to the limited diversity of possible observations in this particular scene. We use the top-$1$ metric for the closed-loop evaluations.

\begin{figure}
    \centering
    \includegraphics[width = \textwidth]{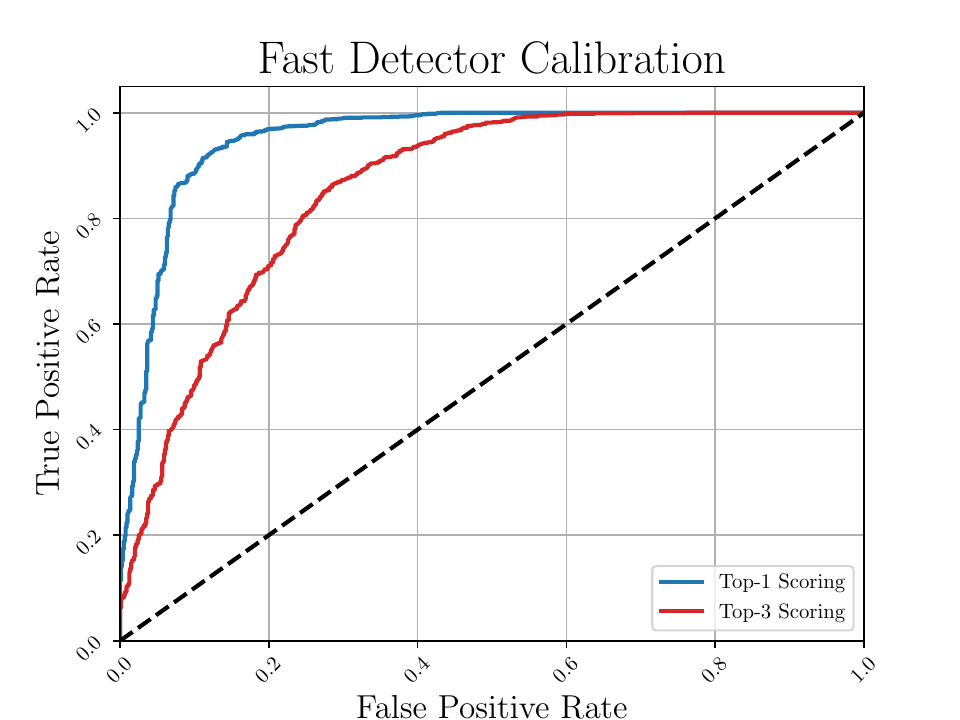}
    \caption{ROC curves computed for the calibration embeddings against the quadrotor's fast anomaly detector's embedding cache.}
    \label{fig:quadrotor-hardware-calibration}
\end{figure}

\subsubsection{Choosing $K$ - Analysis of LLM Query Latencies}

To choose the upper bound $K$ on the latency of the slow generative reasoner used in the MPC \cref{eq:modification-mpc} and Algorithm \ref{alg:safe-mpc}, we perform a simple experiment using the prompt and the observations from the calibration dataset: We query GPT-3.5-turbo (the slow reasoner in our experiment) $N=500$ times and record the response latency. As shown in \cref{fig:gpt-timing}, the response times follow a bimodal distribution with a mean of $3.1\mathrm{s}$ and a standard deviation of $0.85\mathrm{s}$ and a small fraction of outliers. Therefore, we set $K=4.3\mathrm{s}$, corresponding to the $95\%$ quantile of the response times to, to ensure that the LLM returns within our bound $K$ except for rare outlier latencies. We did not experience any instances where the latency was beyond our bound $K$ in our experiments. This is largely because we only sporadically query the LLM once we detect and anomaly, when we truly require the LLM's response.

\begin{figure}
    \centering
    \includegraphics[width=\linewidth]{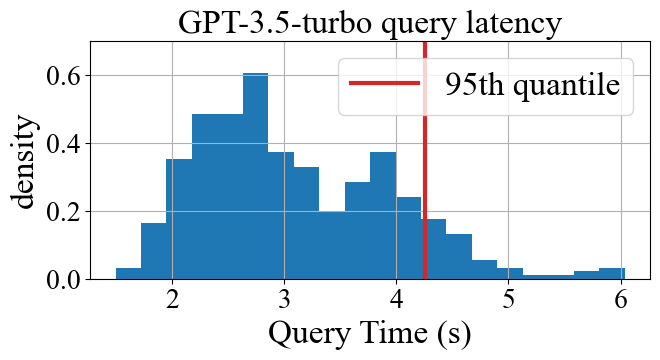}
    \caption{Latency of querying GPT-3.5-Turbo using the hardware experiment's slow reasoner prompt.}
    \label{fig:gpt-timing}
\end{figure}

\subsubsection{Summary of test scenarios}
We describe the qualitative behavior of our scenarios here, but we emphasize to the reader that videos of our experiments are included in the supplementary materials, as well as on the \textbf{project web page: \url{https://sites.google.com/view/aesop-llm}}.  

\textbf{1. Nominal Operation:} There are no obstructions on the red box, so the quadrotor should land normally. As desired, we observe that the quadrotor smoothly flies towards and lands on the box without considering any of the objects on the ground as anomalies.

\textbf{2. Consequential Anomaly:} We consider two variants of this scenario. In the first, another quadrotor has already landed on the red box, necessitating a diversion to the blue box for landing. In the second, quadrotors occupy both red and blue boxes, necessitating a diversion to the holding zone. Qualitatively, these scenarios show three properties of our methodology: First is the ability of both monitors to react to nuanced semantics in the scene from the task context, since ``drones on the ground'' were previously seen in the nominal data but their presence on the landing zones is recognized as safety-critical. Second is the fact that the LLM reasoner helps us select the most appropriate choice of fallback, as it chooses to land on the blue box when possible and recognizes it should recover to the holding zone if not. Third, the fast reasoner ensures that the quadrotor pulls back towards the recovery regions once it detects a hazard on the landing site, rather than naively proceeding with landing while awaiting the LLMs response.

In both these experiments, we see the quadrotor pull back from the red box in the same manner, as the consensus horizon $K$ enforces both recovery plans to be identical untill the LLM returns a decision. We further make a note that the dynamics model used in the MPC does not model ground effect, which results in a significant upward disturbance once the quadrotor attempts its landing on the blue box. In addition, we observed the motion capture system used for state estimation has a dead zone at the location and altitude that the drone reaches above the blue box. As a result, the drone makes a small jump upwards before landing on the blue box.

\textbf{3. Inconsequential Anomaly:} A previously unseen object (specifically, a keyboard) on the ground triggers the fast anomaly detector. However, the subsequent analysis of the slow LLM reasoner correctly deems the anomaly inconsequential, allowing the quadrotor to proceed with landing at its nominal site. In this experiment, we see that once the fast reasoner detects the keyboard, the quadrotor slows down to await the LLM's decision. After the LLM makes its decision the drone speeds back up toward the landing zone.

\subsubsection{Controller Parameters}

To control the quadrotor, we use a Pixracer R15 microcontroller running the open-source PX4 Autopilot software. We use an Optitrack motion capture system for state estimation of the drone, which is fused with the internal IMU of the Pixracer using its built-in EKF. We implement our control stack in ROS2 with nodes written in Python. We implement the MPC controller using a simple kinematic model of the drone, representing the drone's position, attitude, and the rates thereof as the state. Our MPC uses acceleration commands as its inputs, is constrained to maintain the drone's velocity under $1\mathrm{m}/\mathrm{s}$ and within a position/altitude safety fence, and relies on the PX4's internal PID controllers to track desired trajectory setpoints output by the MPC's trajectory predictions. We used a controller horizon of $10\mathrm{s}$ at a time discretization of $0.05\mathrm{s}$ in our experiment, which was sufficiently long to ensure the drone could reach the recovery sets consistently during the experiments. We manually control the drone to liftoff, after which we switch to the MPC controller to execute the trajectory and land the drone. To do so, the MPC nominally controls the drone towards a waypoint a foot above the landing zone. Upon reaching the waypoint, it descends and lands.
}}